\documentclass[journal]{IEEEtran} 
\usepackage{amsmath,amsfonts}
\usepackage[ruled, lined, longend, linesnumbered]{algorithm2e}

\usepackage{bm} 
\usepackage{xcolor}
\usepackage[hidelinks,colorlinks = true,urlcolor=blue,linkcolor = blue,citecolor = blue]{hyperref}
\usepackage{array}
\usepackage{textcomp}
\usepackage{stfloats}
\usepackage{url}
\usepackage{verbatim}
\usepackage{graphicx}
\usepackage{cite}
\usepackage{algpseudocode}
\usepackage{multirow}
\usepackage{arydshln} 
\usepackage{booktabs}

\usepackage{amssymb}
\usepackage{pifont}
\newcommand{\cmark}{\ding{51}}%
\newcommand{\xmark}{\ding{55}}%

\usepackage{float}
\usepackage{tabularx}

\usepackage{subcaption}
\usepackage{enumerate}

\newcommand{\orcidicon}[1]{%
  \href{https://orcid.org/#1}{\includegraphics[width=1em]{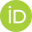}}%
}

\begin{document}
\title{Digital Twin-Driven Communication-Efficient Federated Anomaly Detection for Industrial IoT}

\author{Mohammed Ayalew Belay\textsuperscript{\orcidicon{0000-0003-2144-9790}}, 
Adil Rasheed\textsuperscript{\orcidicon{0000-0003-2690-983X}}, Pierluigi Salvo Rossi\textsuperscript{\orcidicon{0000-0001-6834-8482}}
\thanks{This work has been submitted to the IEEE for possible publication.
Copyright may be transferred without notice, after which this version may no longer be accessible.
}
\thanks{M.A. Belay is with the Department of Electronic Systems, Norwegian University of Science and Technology, 7034 Trondheim, Norway (e-mail: mohammed.a.belay@ntnu.no).} 
\thanks{A. Rasheed is with the Department of Engineering Cybernetics, Norwegian University of Science and Technology, 7034 Trondheim, Norway (e-mail:  adil.rasheed@ntnu.no).}
\thanks{P. Salvo Rossi is with the Department of Electronic Systems, Norwegian University of Science and Technology, 7034 Trondheim, Norway, and with the Department of Gas Technology, SINTEF Energy Research, 7491 Trondheim, Norway (e-mail: salvorossi@ieee.org).}
}




\maketitle

\begin{abstract}

\textcolor{black}{
Anomaly detection is increasingly becoming crucial for maintaining the safety, reliability, and efficiency of industrial systems. Recently, with the advent of digital twins and data-driven decision-making, several statistical and machine-learning methods have been proposed. However, these methods face several challenges, such as dependence on only real sensor datasets, limited labeled data, high false alarm rates, and privacy concerns. To address these problems, we propose a suite of digital twin-integrated federated learning (DTFL) methods that enhance global model performance while preserving data privacy and communication efficiency. Specifically, we present five novel approaches: Digital Twin-Based Meta-Learning (DTML), Federated Parameter Fusion (FPF), Layer-wise Parameter Exchange (LPE), Cyclic Weight Adaptation (CWA), and Digital Twin Knowledge Distillation (DTKD). Each method introduces a unique mechanism to combine synthetic and real-world knowledge, balancing generalization with communication overhead. We conduct an extensive experiment using a publicly available cyber-physical anomaly detection dataset. For a \textbf{target accuracy of 80\%}, CWA reaches the target in \textbf{33}  rounds, FPF in \textbf{41} rounds, LPE in \textbf{48} rounds, and DTML in \textbf{87} rounds, whereas the standard FedAvg baseline and DTKD \textbf{do not reach the target within 100 rounds}. These results highlight substantial communication-efficiency gains (up to \textbf{62\%} fewer rounds than DTML and \textbf{31\%} fewer than LPE) and demonstrate that integrating DT knowledge into FL accelerates convergence to operationally meaningful accuracy thresholds for IIoT anomaly detection.}
\end{abstract}


\section{Introduction}
\label{sec:introduction}

\IEEEPARstart{R}{ecent} advances in Industry 4.0, driven by advanced computing and networked systems, enabled the widespread adoption of digital twins (DTs). Digital twins represent virtual counterparts of physical systems, providing real-time monitoring, simulation, process optimizations, and anomaly detection \cite{Glaessgen2012TheVehicles, Rasheed2020DigitalPerspective, Tao2017DigitalManufacturing, Tao2019DigitalState-of-the-Art, Sisinni2018IndustrialDirections}. Digital twin-based anomaly detection enables the generation and use of large synthetic datasets, avoiding the costly and often impractical collection of real-world failure data \cite{Xu2019ALearning}. 
 In industrial systems, anomaly detection enables asset monitoring, fault identification, predictive maintenance, and performance optimization \cite{Stojanovic2016Big-data-drivenStudy, Yin2014AMonitoring}. Similarly, in cybersecurity, anomaly detection is crucial for network intrusion detection \cite{Bhuyan2014NetworkTools, Garcia-Teodoro2009Anomaly-basedChallenges}. Consequently, various  centralized anomaly detection methods have been proposed in supervised, semi-supervised, and unsupervised learning paradigms \cite{Belay2025AutoregressiveDetection, Belay2025SparseDetection}. Recently, several deep neural network-based anomaly detection methods have been proposed, demonstrating significant improvements in detecting anomalies within high-dimensional and complex datasets \cite{Chandola2009AnomalySurvey, Pang2021DeepReview}. These methods are implemented using different architectures, including recurrent networks \cite{Su2019RobustNetwork, Malhotra2015LongSeries}, convolutional networks \cite{Ren2019Time-seriesMicrosoft, Zhang2019AData}, autoencoders \cite{Zhang2019AData, Zhou2017AnomalyAutoencoders}, generative adversarial networks \cite{Zhang2021UnsupervisedSignals, Zhao2020MultivariateNetwork}, transformers \cite{Belay2024MTAD:Detection, Tuli2022TranAD:Data}, and graph neural networks \cite{Ma2021ALearning}. 

Despite these advancements, both statistical and deep learning-based anomaly detection methods continue to face several critical challenges \cite{Belay2024MultivariateDecomposition, Belay2023UnsupervisedDirections,Tabella2025FailureDivergence,Tabella2025EnhancementModels, Belay2024Self-SupervisedLocalization,Belay2025UnsupervisedPlantsb}. A key limitation is their reliance on real sensor datasets and the scarcity of labeled data, particularly for rare events, which restricts the development of robust models and often leads to high false alarm rates. The lack of diverse data limits the generalization of these models across different scenarios and variations, resulting in suboptimal performance in real-world applications. Moreover, these methods often depend on centralized data processing, which introduces significant privacy and security risks as sensitive data must be transferred and stored in a central location, making it vulnerable to breaches. Additionally, they typically require centralized high-performance computing infrastructure, making them less suitable for real-time, continuous learning scenarios.

A digital twin-based anomaly detection model can be integrated with a real-time model of physical assets in a federated learning mechanism \cite{Belay2025DigitalDetection, Belay2025DigitalIoT}. 
Federated learning (FL) is a decentralized approach to machine learning that allows models to be trained across multiple devices without transferring the raw data to a central server \cite{BrendanMcMahan2016Communication-EfficientData,Li2020FederatedDirectionsb}.
FL reduces the need for high-performance computing infrastructure and reduces the attack surface for potential breaches \cite{Yazdinejad2024AAttacks, Namakshenas2024FederatedThings}. 
It can also support continual learning, which allows models to adapt quickly to new data and emerging trends without waiting for retraining at the central server \cite{Tang2024DigitalDriving}. 

To address the limited availability of labeled data, digital twins are employed to generate synthetic datasets \cite{Qin2024MachineSurvey, Elayan2021DigitalSystems, Gupta2021HierarchicalHealthcare}. Benedictis \textit{et al.} \cite{DeBenedictis2023DigitalProof-of-Concept} introduce a conceptual architecture for industrial Internet of Things (IIoT) anomaly detection based on digital twins and autonomic computing paradigms. Gupta \textit{et al.} \cite{Gupta2023IntegrationThings} introduce a hierarchical federated learning (HFL)-based anomaly detection model for Vehicular Internet of Things (V-IoT). They focus on autonomous vehicles and intelligent transportation systems that involve connected vehicles that communicate with various sensors and IoT devices. Kamalakannan \textit{et al.} \cite{Kamalakannan2024DTFL-DF:Industry} propose DTFL-DF, a digital twin-based federated learning approach, to mitigate fire accidents in the mining industry. They introduce a modified random forest method called Federated Decision Tree, customized for a federated fog environment, to improve fire prediction with low latency. Chatterjee \textit{et al.} \cite{Chatterjee2024DigitalAdvancements} propose blockchain-enabled federated learning (BFL) for credit card fraud detection, merging digital twins with federated learning to create a dynamic approach for identifying known and emerging fraud patterns effectively. Gupta \textit{et al.} \cite{Gupta2021HierarchicalHealthcare} propose a digital twin-based hierarchical federated anomaly detection approach for smart health applications. They develop a federated, time-distributed long short-term memory model to enhance the anomaly detection process.
Praharaj \textit{et al.} \cite{Praharaj2023HierarchicalFarming} propose a similar digital twin-based hierarchical federated anomaly detection approach for smart farming applications. 

Although several digital twin-based federated anomaly detection frameworks have been introduced, several challenges still need to be addressed. One significant challenge is that existing frameworks are purely digital twin-based and do not integrate digital twin models with their physical counterparts effectively. There is a notable absence of methodologies for seamlessly integrating digital twin models with real-world data models. The lack of integration can lead to discrepancies between the virtual and real-world systems. 
The other challenge is communication overhead. Federated learning inherently requires frequent communication between devices and the central server, leading to increased latency and bandwidth consumption. In this study, we propose various digital twin-based anomaly detection frameworks that integrate digital twins with physical systems in a federated learning mechanism. Table~\ref{tab:comparison_FL} highlights the key features of the proposed work compared to some closely related works.

\begin{table*}[t!]
\renewcommand{\arraystretch}{1.5}
\centering
\begin{tabular}{@{}p{5cm}p{1cm}p{1.2cm}p{1.8cm}p{2.5cm}p{2.5cm}@{}}
\toprule
\textbf{Related work} & \textbf{FL} & \textbf{DT Data}  &  \textbf{Physical Asset Integration} & \textbf{Commun.~Overhead Reduction} & \textbf{Learning Approach}\\
\hline
 {Conceptual DT and autonomic computing} \cite{DeBenedictis2023DigitalProof-of-Concept}  & \xmark & \cmark & \xmark & \xmark & Supervised\\
\hline
 {Hierarchical FL for vehicular IoT} \cite{Gupta2023IntegrationThings}  & \cmark & \cmark & \xmark & \xmark & Supervised\\
\hline
 {FL decision forest} \cite{Kamalakannan2024DTFL-DF:Industry}  & \cmark & \cmark & \xmark & \xmark & Supervised\\
\hline
 {Blockchain-enabled FL for fraud detection} \cite{Chatterjee2024DigitalAdvancements}  & \cmark & \cmark & \xmark & \xmark & Supervised\\
\hline
{Hierarchical FL for anomaly detection in smart healthcare} \cite{Gupta2021HierarchicalHealthcare}  & \cmark & \cmark & \xmark & \xmark & Supervised\\
\hline
{Hierarchical federated transfer learning} \cite{Praharaj2023HierarchicalFarming}   & \cmark & \cmark & \xmark & \xmark & Supervised\\
\hline
\textbf{Proposed Methods} & \cmark & \cmark & \cmark & \cmark &  Supervised/Semi-supervised\\
\bottomrule
\end{tabular}
\caption{Overview of methods for digital twin-based and/or federated anomaly detection.}

\label{tab:comparison_FL}
\end{table*}

\color{black}
In this paper, we propose a hybrid and communication-efficient anomaly detection framework employing both DT and FL paradigms to address the aforementioned challenges. The integration of DT and FL into anomaly detection offers the following advantages: 
(i) DTs can generate vast amounts of data, which can be leveraged to train robust anomaly detection models. Moreover, training via synthetic data from DTs and real-world data from multiple physical assets enables better generalization of the anomaly detection model.
(ii) FL enhances the training  process by allowing these models to be trained across multiple physical assets without compromising data privacy. The proposed methods significantly reduce communication overhead during FL rounds, ensuring scalability and computational efficiency. Specifically, the contributions of the paper are summarized as follows:
\begin{itemize}
    \item We propose four supervised and one semi-supervised hybrid learning mechanism for a federated and digital twin-based anomaly detection.

    \item We proposed a digital twin-based knowledge distillation, and we provide a detailed computational complexity analysis of the proposed methods. 
    
    \item We performed an extensive performance analysis using publicly available datasets from real-world digital and physical assets.
\end{itemize}

The rest of the paper is structured as follows: 
Section~\ref{sec:methodology} describes the proposed method;
Section~\ref{sec:Experiments} presents the experimental setup and the datasets;
Section~\ref{sec:Results} illustrates the results from the performance analysis and includes the related discussion; 
finally, conclusions and future research directions are given in Section~\ref{sec:Conclusions}.

\emph{Notation} -- Vectors and matrices are denoted by bold lower-case and upper-case letters, respectively. 

\color{black}

\section{The Proposed Methods}
\label{sec:methodology}


\begin{table*}[t!]
\centering
\renewcommand{\arraystretch}{1.5}
\begin{tabular}{lll}
\hline
\textbf{Method} & \textbf{Update Rule} & \textbf{Learning / Training}\\
\hline
DTML & FedAvg with digital-twin meta-gradient refinement & Supervised / Hybrid\\
FPF  & Weighted fusion of client and DT parameters & Supervised / Hybrid\\
LPE  & Layer-level bidirectional parameter overwriting & Supervised / Hybrid\\
CWA  & Alternating overwrite of parameters & Supervised / Hybrid\\
DTKD & Teacher–student distillation via soft labels & Semi-supervised / DT-pretrained\\
\hline
\end{tabular}
\caption{Summary of proposed approaches with their respective update rules and learning/training approaches.}
\label{tab:proposed_approaches}
\end{table*}


We propose a hybrid digital twin-based federated learning framework for anomaly detection in industrial IoT (IIoT). The main objective is to collaboratively train a robust global anomaly detection model leveraging both simulated data from digital twins and real-world data from distributed physical assets. This process aims at preserving data privacy, enhancing model robustness, and minimizing communication overhead. The key components and processes of the proposed framework include the following:
\begin{itemize}
    \item Multiple physical systems equipped with local datasets containing operational data possibly related to both normal and anomalous events;
    \item A DT producing a synthetic dataset simulating a wide range of operational scenarios for the physical systems;    
    \item A global model for anomaly detection trained using both synthetic data from the DT and local data from the physical systems; 
    \item The training is based on the different aggregation of local models using FL and    
\end{itemize}

\subsection{DT-based Federated Learning}
\textcolor{black}{The proposed framework involves distributed optimization of a global objective across multiple assets.} 
More formally, we consider $K$ physical assets, where $\mathcal{D}_k=\{(\bm{x}_{k,i}, y_{k,i})\}_{i=1}^{n_k}$ denotes the local dataset of the $k$th asset, $n_k$ the number of samples, $\bm{x}_{k,i}\in\mathbb{R}^d$ is the feature vector for the $i$th sample, and $y_{k,i}\in\{0,1\}$ is the corresponding label indicating whether the sample is normal ($y_{k,i}=0$) or anomalous ($y_{k,i}=1$). In a federated learning framework, the objective is to collaboratively optimize a global anomaly detection model parameterized by $\bm{\Theta}$ that minimizes:
\begin{equation}
\bm{\Theta}^* = \arg\min_{\bm{\Theta}} \frac{1}{\sum_{k=1}^{K} n_k} \sum_{k=1}^{K} \sum_{j=1}^{n_k} \ell_\text{BCE}(\bm{\Theta}; d_{k,j}),
\label{eq:global_objective}
\end{equation}
where $\ell_\text{phys}(\bm{\Theta}; d_{k,j})$ is the binary cross entropy local loss for the $j$-th data sample $d_{k,j}$ at asset $k$. The FL training proceeds in iterative rounds ($t = 1,2,\dots$). At each iteration, the central server randomly selects a subset $S_t$ of assets, with size: $m = \max(C \cdot K, 1)$, where $C$ represents the fraction of participating clients. 
Each client $k$ splits its dataset $\mathcal{D}_k$ into batches of size $B$, and performs a gradient descent step for several epochs ($E$) via mini-batch stochastic gradient descent:
\begin{align}
\bm{\Theta}^{(t+1)}_{k} = \bm{\Theta}^{(t)} -  \frac{\eta_{k}^{(t)}}{n^{(t)}_{k}} \nabla_{\bm{\Theta}} \sum_{d_{k,j} \in \mathcal{D}^{(t)}_{k}} \ell(\bm{\Theta}^{(t)}; d_{k,j}) 
\label{eq:local_gradient}
\end{align}
where $\eta_k^{(t)}$ is the local learning rate and $\mathcal{D}^{(t)}_k$ is a randomly selected batch of data.
In a standard federated learning, the central server aggregates these local parameters by averaging \cite{McMahan2017Communication-EfficientData}:
\begin{align}
 \bm{\Theta}^{(t+1)} &= \frac{1}{K} \sum_{k=1}^{K} \bm{\Theta}^{(t+1)}_{k}.
 \label{eq:model_aggregation}
\end{align}
This iterative global model update continues until convergence. 

In a digital twin-based federated learning, a digital twin provides a synthetic dataset $\mathcal{D}_{\text{twin}}$ that captures a broad range of operational conditions. A standalone DT-based model, denoted $\bm{\Theta}_{\text{twin}}$, is trained using this dataset and shared with all physical systems. We introduce a hybrid learning mechanism that integrates parameters from both synthetic (DT) and real-world sources. In this setup:
\begin{itemize}
    \item The DT model resides on a central server and serves as a global prior;
    \item Each physical client trains a local model using both its own dataset $\mathcal{D}_k$ and the knowledge from $\bm{\Theta}_{\text{twin}}$;
    \item Gradients or model updates are integrated through a variety of strategies to align and enhance learning.
\end{itemize}

Five methods are proposed to integrate DT and client parameters during training:
\begin{enumerate}

    \item \textbf{DT-based Meta-Learning (DTML)}: Learns a global initialization using client adaptation and refines it using digital twin data.

    \item \textbf{Federated Parameter Fusion (FPF)}: Aggregates client parameters using similarity-weighted averaging with the digital twin to form a robust global model.

    \item \textbf{Layer-wise Parameter Exchange (LPE)}: Selectively replaces layers between the digital twin and the aggregated model to enable fine-grained bidirectional knowledge transfer, using static or similarity-based layer selection policies.
    
    \item \textbf{Cyclic Weight Adaptation (CWA)}: Alternates updates between client-aggregated parameters and the digital twin to synchronize learning dynamics.
    
    \item \textbf{DT Knowledge Distillation (DTKD)}: Uses soft labels from the digital twin to guide client learning via KL-divergence, enabling semi-supervised training without ground-truth labels.

\end{enumerate}
A summary of the proposed approaches and their attributes is presented in Table \ref{tab:proposed_approaches}.


\subsection{DT-based Meta-Learning}

\begin{figure}[t!]\centering
\includegraphics[width=1.0\columnwidth]{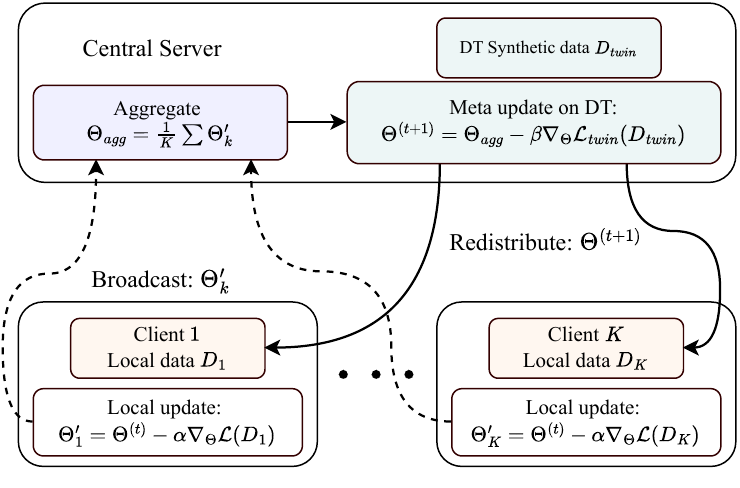}
\caption{\textcolor{black}{DT-based meta-learning framework.}}
\label{fig:DTML}
\end{figure}

\textcolor{black}{We propose a hybrid federated meta-learning framework that leverages a \textit{digital twin} as a global validator to improve generalization in federated anomaly detection, as shown in Figure \ref{fig:DTML}.} Unlike standard federated averaging, which simply aggregates client models, MLDT provides a global initialization that enables each client to rapidly adapt to its local task while ensuring the aggregated model generalizes well on digital twin data. Let $ \bm{\Theta}^{(t)} $ be the global model at round $ t $, and $ \mathcal{D}_k $ denote the training set for client $ k $. Each client performs a local update initialized from the shared model:
\begin{equation}
\bm{\Theta}_k' = \bm{\Theta}^{(t)} - \alpha \nabla_{\bm{\Theta}} \sum_{(\mathbf{x}, y) \in \mathcal{D}_k} \ell(\bm{\Theta}; \mathbf{x}, y)
\end{equation}
where $ \alpha $ is the local learning rate. After local adaptation, the server aggregates the updated parameters. However, instead of directly using aggregation as the new global model, MLDT performs a meta-update using digital twin validation data $ \mathcal{D}_{\text{twin}} $:
\begin{equation}
\bm{\Theta}^{(t+1)} = \left( \frac{1}{K} \sum_{k=1}^K \bm{\Theta}_k' \right)  - \beta \nabla_{\bm{\Theta}} \mathcal{L}_{\text{twin}}(\bm{\Theta}; \mathcal{D}_{\text{twin}}) \bigg|_{\bm{\Theta} = \bm{\Theta}_{\text{agg}}^{(t+1)}}
\end{equation}
where $ \beta $ is the outer-loop or meta-learning rate and $ \mathcal{L}_{\text{twin}} $ is the binary cross-entropy loss on the synthetic data:
\begin{equation}
\mathcal{L}_{\text{twin}}(\bm{\Theta}; \mathcal{D}_{\text{twin}}) = \frac{1}{|\mathcal{D}_{\text{twin}}|} \sum_{(\mathbf{x}, y) \in \mathcal{D}_{\text{twin}}} \ell_{\text{BCE}}(\bm{\Theta}; \mathbf{x}, y)
\end{equation}
This update ensures that the global model not only captures aggregated client knowledge but also generalizes well under the operational conditions simulated by the digital twin. The refined global model $\bm{\Theta}^{(t+1)}$ is then redistributed to all clients for the next training round. 
\color{black}
In the above formulation, we assume that client datasets $\{\mathcal{D}_k\}$ are independently and identically distributed (IID), which simplifies aggregation and meta-updates. However, in real-world IIoT systems, client data are often \textit{non-IID}, leading to significant heterogeneity across clients. Under such conditions, local model updates can diverge, and in the case of DTML, meta-gradients may amplify these divergences, causing instability during training. The pseudocode for DTML is presented in Algorithm \ref{alg:algorithm_1}. 

\color{black}

\begin{algorithm}[t]
\SetKwInOut{Input}{\textcolor{black}{Input}}
\SetKwInOut{Output}{\textcolor{black}{Output}}
\caption{\textcolor{black}{DT-based meta-learning (DTML)}}
\Input{\textcolor{black}{Global model $\Theta^{(0)}$, DT model $\Theta_{\mathrm{twin}}^{(0)}$, client datasets $\{D_k\}_{k=1}^K$, rounds $T$, client fraction $C$, local epochs $E$, batch size $B$, local LR $\eta$, meta LR $\beta$, twin data $D_{\mathrm{twin}}$}}
\Output{\textcolor{black}{Final global model $\Theta^{(T)}$}}

\For{\textcolor{black}{$t=0,1,\dots,T-1$}}{
  \textcolor{black}{Server selects $S_t \subset \{1,\ldots,K\}$ with $|S_t| = m=\max(1,\lfloor C K \rfloor)$}\;
  \textcolor{black}{Server broadcasts $\Theta^{(t)}$ to clients in $S_t$}\;

  \tcp{\textcolor{black}{Local adaptation}}
  \ForEach{\textcolor{black}{client $k \in S_t$ \textbf{in parallel}}}{
    \textcolor{black}{$\Theta_k \leftarrow \Theta^{(t)}$}\;
    \For{\textcolor{black}{$e=1$ \KwTo $E$}}{
      \For{\textcolor{black}{mini-batch $\mathcal{B} \subset D_k$ of size $B$}}{
        \textcolor{black}{$\Theta_k \leftarrow \Theta_k - \eta \nabla_\Theta \ell_{\mathrm{BCE}}(\Theta_k;\mathcal{B})$}\;
      }
    }
    \textcolor{black}{Send adapted parameters $\Theta'_k \leftarrow \Theta_k$ to server}\;
  }
  \tcp{\textcolor{black}{Aggregation then meta-update on twin data}}
  \textcolor{black}{$\Theta_{\mathrm{agg}} \leftarrow \frac{1}{|S_t|}\sum_{k \in S_t} \Theta'_k$}\;
  \textcolor{black}{$\Theta^{(t+1)} \leftarrow \Theta_{\mathrm{agg}} - \beta \,\nabla_\Theta L_{\mathrm{twin}}(\Theta; D_{\mathrm{twin}})\big|_{\Theta=\Theta_{\mathrm{agg}}}$}\;
  
}
\label{alg:algorithm_1}
\end{algorithm}

\subsection{DT-based Federated Parameter Fusion}

\begin{figure}[t!]\centering
\includegraphics[width=1.0\columnwidth]{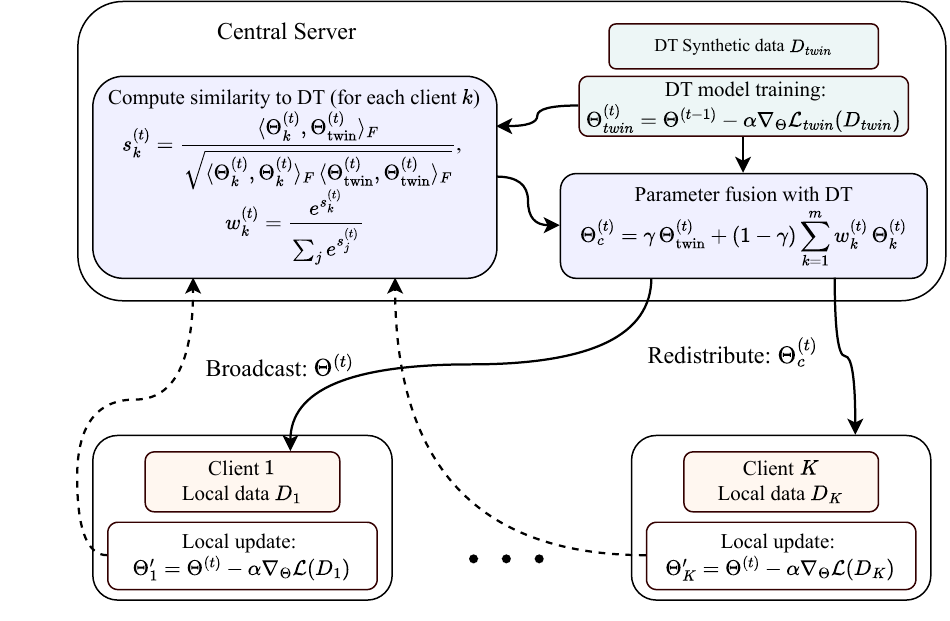}
\caption{\textcolor{black}{DT-based federated parameter fusion framework.}}
\label{fig:FPF}
\end{figure}

\textcolor{black}{In the parameter fusion approach, the server aggregates the client model parameters using federated averaging and adaptively blends them with the digital twin model parameters to update both the digital twin model and the global model, as shown in Figure \ref{fig:FPF}.} Moreover, the method favors updates from clients whose models are more aligned with the digital twin. Clients that are better aligned with the DT likely represent more reliable or relevant operational scenarios. Let $\bm{\Theta}_k^{(t)}$ be the model parameters from the client $k$ at round $t$, and $\bm{\Theta}_{\text{twin}}^{(t)}$ be the digital twin model. We compute a similarity score $s_k^{(t)}$ between the client and twin model using RV coefficient (vector correlation) as:
\begin{equation}
s_k^{(t)} = \frac{\langle \bm{\Theta}_k^{(t)}, \bm{\Theta}_{\text{twin}}^{(t)} \rangle_F}{\sqrt{\langle \bm{\Theta}_k^{(t)}, \bm{\Theta}_k^{(t)} \rangle_F \langle \bm{\Theta}_{\text{twin}}^{(t)}, \bm{\Theta}_{\text{twin}}^{(t)} \rangle_F}}
\end{equation}
where $\langle \cdot, \cdot \rangle_F$ denotes the Frobenius inner product.
We normalize the client scores using a softmax function to ensure non-negativity and that weights sum to 1:

\begin{equation}
w_k^{(t)} = \frac{\exp(s_k^{(t)})}{\sum_{j=1}^{K} \exp(s_j^{(t)})} \;.
\end{equation}
The global model update is then computed as the weighted sum of the client models and the digital twin parameters as:
\begin{equation}
\bm{\Theta}_{\text{c}}^{(t)} = \gamma \bm{\Theta}_{\text{twin}}^{(t)} + (1 - \gamma) \sum_{k=1}^{K} w_k^t \bm{\Theta}_k^{(t)} \;.
\end{equation}
where $\gamma \in [0, 1]$ is a weighting factor and controls the contribution of the digital twin model and client models. The fusion ensures that both the global model and local models incorporate knowledge from both real-world and synthetic data, providing a balanced and robust learning framework. Next, the combined parameters update the digital twin and are also sent back to the clients to update their models as:
\begin{equation}
\bm{\Theta}_{\text{twin}}^{(t+1)} \leftarrow \bm{\Theta}_{\text{c}}^{(t)}, \quad \bm{\Theta}_{k}^{(t+1)} \leftarrow \bm{\Theta}_{\text{c}}^{(t)}
\end{equation}
This approach dynamically emphasizes clients with stronger similarity to the digital twin model, leading to a robust and domain-aligned aggregation strategy. Furthermore, it reduces the influence of outliers or noisy clients, which is crucial in industrial IoT systems with diverse operational conditions. The pseudocode for FPF is summarized in Algorithm \ref{alg:algorithm_2}. 

\begin{algorithm}[t]
\SetKwInOut{Input}{\textcolor{black}{Input}}
\SetKwInOut{Output}{\textcolor{black}{Output}}
\caption{\textcolor{black}{Federated Parameter Fusion (FPF)}}
\Input{\textcolor{black}{Global $\Theta^{(0)}$, DT $\Theta_{\mathrm{twin}}^{(0)}$, $\{D_k\}$, rounds $T$, $C,E,B,\eta$, fusion weight $\gamma \in [0,1]$, similarity function $\mathrm{sim}(\cdot,\cdot)$ (default: Frobenius/RV as in Eq.~(7))}}
\Output{\textcolor{black}{Final global $\Theta^{(T)}$, updated DT $\Theta_{\mathrm{twin}}^{(T)}$}}

\For{\textcolor{black}{$t=0,1,\dots,T-1$}}{
  \textcolor{black}{Select clients $S_t$ and broadcast $\Theta^{(t)}$}\;
  \ForEach{\textcolor{black}{$k \in S_t$ \textbf{in parallel}}}{
    \textcolor{black}{Local train $\Theta_k^{(t)} \leftarrow \text{SGD}(\Theta^{(t)}; D_k, E,B,\eta)$}\;
    \textcolor{black}{Send $\Theta_k^{(t)}$ to server}\;
  }

  \tcp{\textcolor{black}{Compute similarity weights w.r.t. current twin}}
  \ForEach{\textcolor{black}{$k \in S_t$}}{
    \textcolor{black}{$s_k \leftarrow \mathrm{sim}\!\left(\Theta_k^{(t)}, \Theta_{\mathrm{twin}}^{(t)}\right)$}\;
  }
  \textcolor{black}{$w_k \leftarrow \frac{\exp(s_k)}{\sum_{j \in S_t}\exp(s_j)}$ for all $k \in S_t$}\;

  \tcp{\textcolor{black}{Fuse client and twin parameters (Eq.~(9))}}
  \textcolor{black}{$\Theta_c \leftarrow \gamma \,\Theta_{\mathrm{twin}}^{(t)} + (1-\gamma)\sum_{k \in S_t} w_k \Theta_k^{(t)}$}\;

  \tcp{\textcolor{black}{Update twin and broadcast fused model}}
  \textcolor{black}{$\Theta_{\mathrm{twin}}^{(t+1)} \leftarrow \Theta_c$}\;
  \textcolor{black}{$\Theta^{(t+1)} \leftarrow \Theta_c$ and broadcast to all clients (or $S_t$)}\;
}
\label{alg:algorithm_2}
\end{algorithm}


\subsection{Layer-wise Parameter Exchange}

\begin{figure}[t!]\centering
\includegraphics[width=1.0\columnwidth]{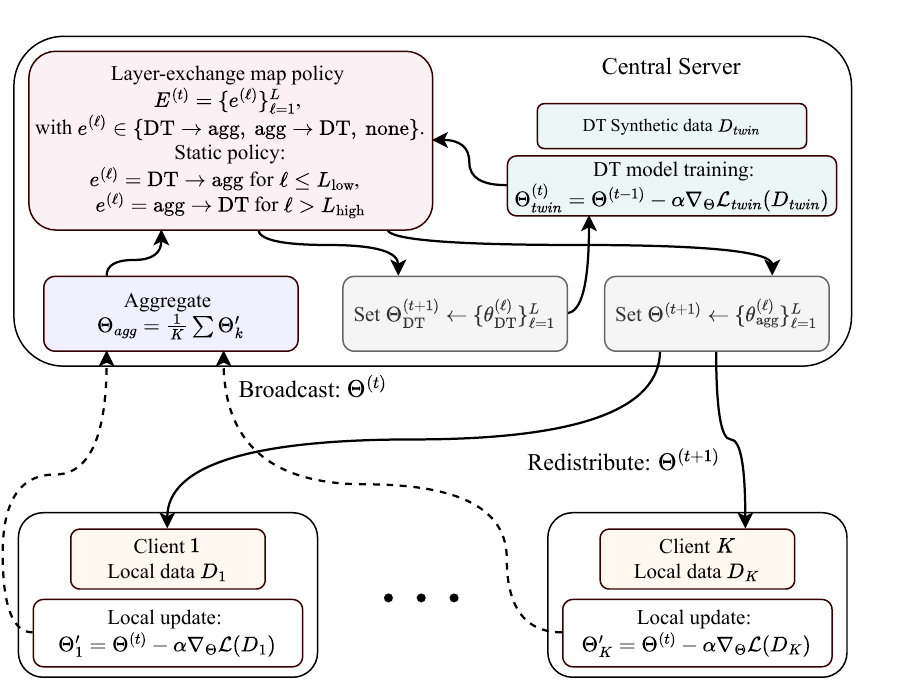}
\caption{\textcolor{black}{Layer-wise parameter exchange framework.}}
\label{fig:LPE}
\end{figure}

\begin{figure}[t!]\centering
\includegraphics[width=0.95\columnwidth]{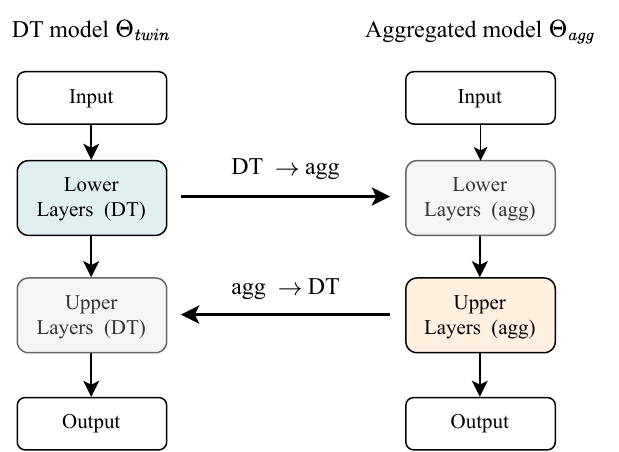}
\caption{\textcolor{black}{Training flow with layer-exchange map \(\mathcal{E}^{(t)}\) that controls bidirectional, layer-granular transfer between the DT and the aggregated global model. With a static policy proposed: lower layers copied DT\(\rightarrow\)agg; upper layers copied agg\(\rightarrow\)DT.}}
\label{fig:LPE2}
\end{figure}

While prior approaches fuse or align entire model parameters, \textit{layer-wise exchange} shown in Figure \ref{fig:LPE} enables selective updates between the digital twin and aggregated models at the granularity of network layers.
This supports fine-grained knowledge transfer, domain-specific adaptation, and better generalization across diverse IIoT environments. Let the global model be composed of $L$ layers with parameters:
\begin{equation}
\bm{\Theta} = \left\{\bm{\theta}^{(1)}, \bm{\theta}^{(2)}, \dots, \bm{\theta}^{(L)}\right\}
\end{equation}
where $\bm{\theta}^{(\ell)}$ are the parameters of the $\ell$-th layer. We define a \textit{layer exchange map}:
\begin{equation}
\mathcal{E}^{(t)} = \left\{e^{(\ell)} \in \{\text{DT} \rightarrow \text{agg},\ \text{agg} \rightarrow \text{DT},\ \text{none}\} \right\}_{\ell=1}^L
\end{equation}

which controls the direction of layer-wise parameter exchange. After client aggregation, the server updates each layer as:
\begin{equation}
\bm{\theta}^{(\ell)}_{\text{DT}} \leftarrow 
\begin{cases}
\bm{\theta}^{(\ell)}_{\text{agg}}, & \text{if } e^{(\ell)} = \text{agg} \rightarrow \text{DT} \\
\bm{\theta}^{(\ell)}_{\text{DT}}, & \text{otherwise}
\end{cases}
\end{equation}

\begin{equation}
\bm{\theta}^{(\ell)}_{\text{agg}} \leftarrow 
\begin{cases}
\bm{\theta}^{(\ell)}_{\text{DT}}, & \text{if } e^{(\ell)} = \text{DT} \rightarrow \text{agg} \\
\bm{\theta}^{(\ell)}_{\text{agg}}, & \text{otherwise}
\end{cases}
\end{equation}

The updated digital twin and global model are:
\begin{equation}
\bm{\Theta}_{\text{DT}}^{(t+1)} = \{\bm{\theta}^{(\ell)}_{\text{DT}}\}_{\ell=1}^L, \quad
\bm{\Theta}^{(t+1)} = \{\bm{\theta}^{(\ell)}_{\text{agg}}\}_{\ell=1}^L
\end{equation}

We specify a static exchange policy (lower layers from DT, upper layers from clients), and the exchange rule $e^{(\ell)}$ is given by:
\begin{equation}
e^{(\ell)} =
\begin{cases}
\text{DT} \rightarrow \text{agg}, & \ell \leq L_{\text{low}} \\
\text{agg} \rightarrow \text{DT}, & \ell > L_{\text{high}} \\
\text{none}, & \text{otherwise}
\end{cases}
\end{equation}
Clients receive the updated model $\bm{\Theta}^{(t+1)}$ and perform standard local updates:
\begin{equation}
\bm{\Theta}_k^{(t+1)} = \bm{\Theta}^{(t+1)} - \eta \nabla \mathcal{L}_{\text{BCE}}(\bm{\Theta}^{(t+1)}; \mathcal{D}_k)
\end{equation}
Layer-wise parameter exchange provides a flexible mechanism for hybrid learning, enabling domain-specific knowledge sharing at different network depths and fine-grained control of transfer between DT and client knowledge. The pseudocode for FPF is summarized in Algorithm \ref{alg:algorithm_3}. 

\begin{algorithm}[t]
\SetKwInOut{Input}{\textcolor{black}{Input}}
\SetKwInOut{Output}{\textcolor{black}{Output}}
\caption{\textcolor{black}{Layer-wise Parameter Exchange (LPE)}}
\Input{\textcolor{black}{Layered models $\Theta=\{\theta^{(1)},\dots,\theta^{(L)}\}$, $\Theta_{\mathrm{twin}}=\{\theta_{\mathrm{twin}}^{(1)},\dots,\theta_{\mathrm{twin}}^{(L)}\}$, policy $E^{(t)}=\{e^{(\ell)}\}$ with $e^{(\ell)} \in \{\mathrm{DT}\!\to\!\mathrm{agg},\,\mathrm{agg}\!\to\!\mathrm{DT},\,\mathrm{none}\}$ (Eq.~(16)), $C,E,B,\eta$}}
\Output{\textcolor{black}{Updated $\Theta^{(T)}$, $\Theta_{\mathrm{twin}}^{(T)}$}}

\For{\textcolor{black}{$t=0,1,\dots,T-1$}}{
  \textcolor{black}{Select $S_t$ and broadcast $\Theta^{(t)}$}\;
  \ForEach{\textcolor{black}{$k \in S_t$ \textbf{in parallel}}}{
    \textcolor{black}{$\Theta_k^{(t)} \leftarrow \text{SGD}(\Theta^{(t)}; D_k, E,B,\eta)$}\;
    \textcolor{black}{Send $\Theta_k^{(t)}$ to server}\;
  }
  \textcolor{black}{$\Theta_{\mathrm{agg}} \leftarrow \frac{1}{|S_t|}\sum_{k\in S_t}\Theta_k^{(t)}$}\;

  \tcp{\textcolor{black}{Layer-wise exchange between $\Theta_{\mathrm{agg}}$ and $\Theta_{\mathrm{twin}}^{(t)}$}}
  \For{\textcolor{black}{$\ell=1$ \KwTo $L$}}{
    \uIf{\textcolor{black}{$e^{(\ell)}=\mathrm{DT}\!\to\!\mathrm{agg}$}}{
      \textcolor{black}{$\theta_{\mathrm{agg}}^{(\ell)} \leftarrow \theta_{\mathrm{twin}}^{(\ell)}$}\;
    }\uElseIf{\textcolor{black}{$e^{(\ell)}=\mathrm{agg}\!\to\!\mathrm{DT}$}}{
      \textcolor{black}{$\theta_{\mathrm{twin}}^{(\ell)} \leftarrow \theta_{\mathrm{agg}}^{(\ell)}$}\;
    }
  }

  \tcp{\textcolor{black}{Finalize and broadcast}}
  \textcolor{black}{$\Theta^{(t+1)} \leftarrow \Theta_{\mathrm{agg}}$; $\Theta_{\mathrm{twin}}^{(t+1)} \leftarrow \Theta_{\mathrm{twin}}$}\;
  \textcolor{black}{Broadcast either full $\Theta^{(t+1)}$ or only changed layers (comm-efficient variant)}\;
}
\label{alg:algorithm_3}
\end{algorithm}


\subsection{Cyclic Weight Adaptation}
\begin{figure}[t!]\centering
\includegraphics[width=1.0\columnwidth]{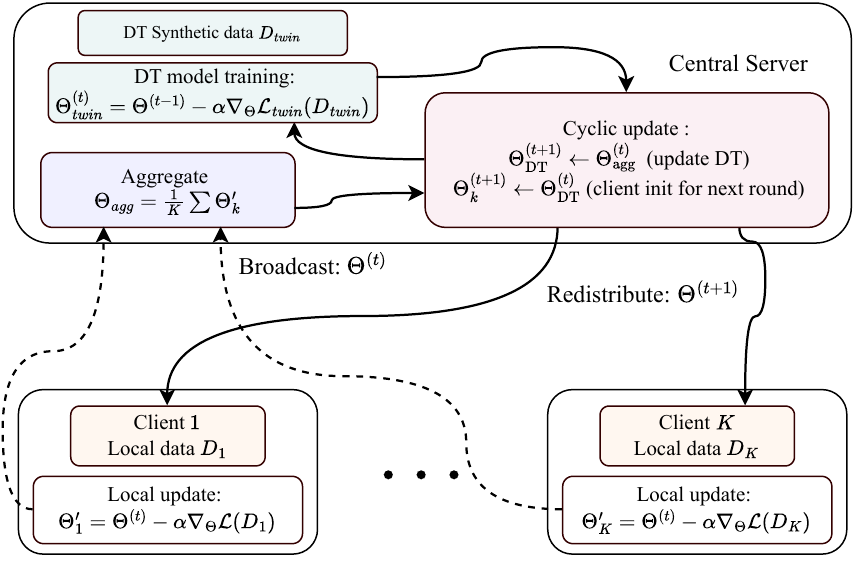}
\caption{\textcolor{black}{Cyclic weight adaptation framework.}}
\label{fig:CWA}
\end{figure}

In cyclic weight adaptation, we alternate between DT and client parameters in successive updates, as illustrated in Figure \ref{fig:CWA}. After standard federated averaging, we perform cyclic updates:
\begin{equation}
\bm{\Theta}_{\text{DT}}^{(t+1)} \leftarrow \bm{\Theta}_{\text{agg}}^{(t)}, \quad \bm{\Theta}_k^{(t+1)} \leftarrow \bm{\Theta}_{\text{DT}}^{(t)}
\end{equation}
This enforces synchronized progression, alternating directional influence between synthetic and real models. The pseudocode for CWA is summarized in Algorithm \ref{alg:algorithm_4}. 

\begin{algorithm}[t]
\SetKwInOut{Input}{\textcolor{black}{Input}}
\SetKwInOut{Output}{\textcolor{black}{Output}}
\caption{\textcolor{black}{Cyclic Weight Adaptation (CWA)}}
\Input{\textcolor{black}{Global $\Theta^{(0)}$, DT $\Theta_{\mathrm{twin}}^{(0)}$, $\{D_k\}$, rounds $T$, $C,E,B,\eta$}}
\Output{\textcolor{black}{Final $\Theta^{(T)}$, $\Theta_{\mathrm{twin}}^{(T)}$}}

\For{\textcolor{black}{$t=0,1,\dots,T-1$}}{
  \textcolor{black}{Select $S_t$ and broadcast $\Theta^{(t)}$}\;
  \ForEach{\textcolor{black}{$k \in S_t$ \textbf{in parallel}}}{
    \textcolor{black}{$\Theta_k^{(t)} \leftarrow \text{SGD}(\Theta^{(t)}; D_k, E,B,\eta)$}\;
    \textcolor{black}{Send $\Theta_k^{(t)}$ to server}\;
  }
  \textcolor{black}{$\Theta_{\mathrm{agg}} \leftarrow \frac{1}{|S_t|}\sum_{k\in S_t}\Theta_k^{(t)}$}\;

  \tcp{\textcolor{black}{Cycle: alternate influence between DT and clients}}
  \If{\textcolor{black}{$t$ is even}}{
    \textcolor{black}{$\Theta_{\mathrm{twin}}^{(t+1)} \leftarrow \Theta_{\mathrm{agg}}$;\quad $\Theta^{(t+1)} \leftarrow \Theta_{\mathrm{agg}}$}\;
  }\Else{
    \textcolor{black}{$\Theta^{(t+1)} \leftarrow \Theta_{\mathrm{twin}}^{(t)}$;\quad broadcast $\Theta^{(t+1)}$}\;
  }
}
\label{alg:algorithm_4}
\end{algorithm}

\subsection{DT Knowledge Distillation}

\begin{figure}[t!]\centering
\includegraphics[width=1.0\columnwidth]{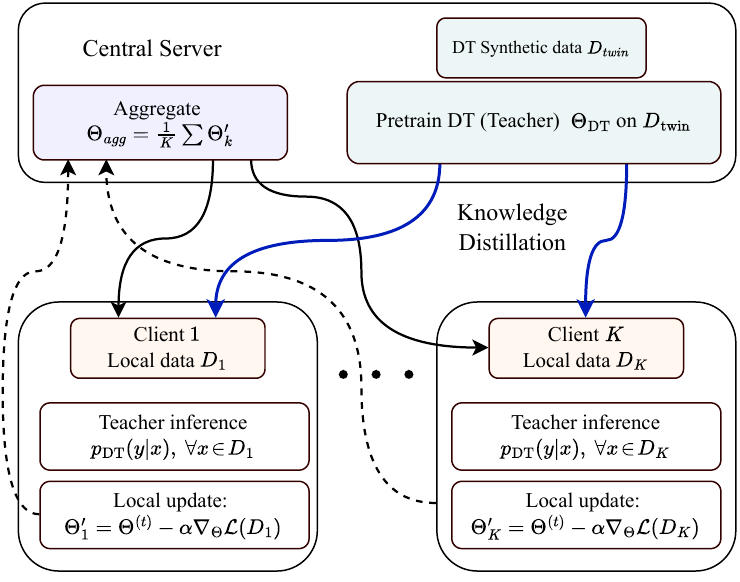}
\caption{\textcolor{black}{DT knowledge distillation framework.}}
\label{fig:DTKD}
\end{figure}

We adopt a semi-supervised framework in which clients learn from soft targets produced by a DT-trained model. First, the global model with parameters is trained on synthetic data. 
The DT produces soft labels $p_{\text{DT}}(y|\mathbf{x}_i)$, which are distilled into client models via Kullback–Leibler (KL) divergence:
\begin{equation}
\ell_{\text{KL}} = D_{\text{KL}}\left(p_{\text{DT}}(y|\mathbf{x}_i) \,\|\, p_k(y|\mathbf{x}_i)\right)
\end{equation}
The KL-divergence loss for the client $k$ over its dataset $\mathcal{D}_k$ is:
\begin{equation}
\mathcal{L}_{\text{KD}} = \frac{1}{|\mathcal{D}_k|}\sum_{\mathbf{x}\in\mathcal{D}_k}\sum_{y\in\{0,1\}} p_{\text{DT}}(y|\mathbf{x})\, \log\!\frac{p_{\text{DT}}(y|\mathbf{x})}{p_k(y|\mathbf{x})}
\end{equation}
This enables knowledge transfer from labeled synthetic data to unlabeled real data while maintaining client privacy.
\color{black}
Although DTKD enables effective transfer of synthetic knowledge, deploying the full teacher model on edge devices may introduce memory and latency overheads. In particular, limited device capacity can restrict storage of large DT models, and repeated teacher inferences for generating soft labels may slow down training. To mitigate this, lightweight strategies such as model quantization, pruning, or using a compact surrogate teacher can reduce resource demands.

\color{black}

\begin{algorithm}[t]
\SetKwInOut{Input}{\textcolor{black}{Input}}
\SetKwInOut{Output}{\textcolor{black}{Output}}
\caption{\textcolor{black}{DT Knowledge Distillation (DTKD)}}
\Input{\textcolor{black}{DT teacher $\Theta_{\mathrm{twin}}$ pretrained on $D_{\mathrm{twin}}$, rounds $T$, $C,E,B,\eta$}} 
\Output{\textcolor{black}{Final student $\Theta^{(T)}$}}

\For{\textcolor{black}{$t=0,1,\dots,T-1$}}{
  \textcolor{black}{Select $S_t$ and broadcast current student $\Theta^{(t)}$ and fixed teacher $\Theta_{\mathrm{twin}}$}\;
  \ForEach{\textcolor{black}{$k \in S_t$ \textbf{in parallel}}}{
    \For{\textcolor{black}{$e=1$ \KwTo $E$}}{
      \For{\textcolor{black}{mini-batch $\mathcal{B} \subset D_k$}}{
        \textcolor{black}{Compute teacher soft targets $p_{\mathrm{t}}(y|x)$}\;
        \textcolor{black}{Compute student logits $p_{\mathrm{s}}(y|x)$}\;
        \textcolor{black}{$\mathcal{L}_{\mathrm{KL}} \leftarrow \frac{1}{|\mathcal{B}|}\sum_{x\in\mathcal{B}} D_{\mathrm{KL}}\!\left(p_{\mathrm{t}}(\cdot|x)\,\|\,p_{\mathrm{s}}(\cdot|x)\right)$}\;
        \textcolor{black}{$\Theta^{(t)}_k \leftarrow \Theta^{(t)}_k - \eta \nabla_\Theta \mathcal{L}_{KL}$}\;
      }
    }
    \textcolor{black}{Send $\Theta^{(t)}_k$ to server}\;
  }
  \tcp{\textcolor{black}{Aggregate students}}
  \textcolor{black}{$\Theta^{(t+1)} \leftarrow \frac{1}{|S_t|}\sum_{k\in S_t}\Theta^{(t)}_k$}\;
}
\label{alg:algorithm_5}
\end{algorithm}

\section{Experimental Setup}

\label{sec:Experiments}
\subsection{Datasets} 
\begin{figure}[t!]\centering
\includegraphics[width=1.0\columnwidth]{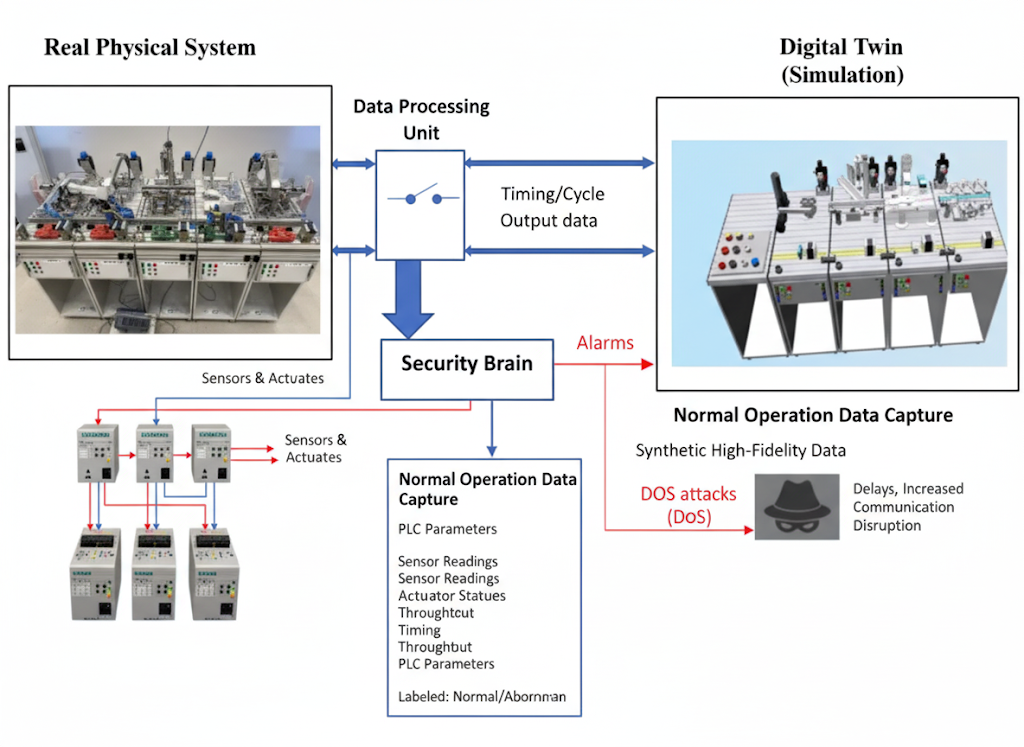}
\caption{\textcolor{black}{Experimental setup for the Industry 4.0 production line system and its digital twin under cyberattack (from \cite{Shi2023AAttack}).}}
\label{fig:DT_data}
\end{figure}

\textcolor{black}{
We evaluate the proposed methods using a dataset from an Industry 4.0 (I4.0) production line system and its digital twin under cyberattack \cite{Shi2023AAttack}. 
The data is collected from both the real manufacturing system and its digital twin when subjected to a denial of service (DoS) attack that creates delays in the system response and increases the cycle time. More specifically, $4$ modular production stations are present and controlled by Siemens programmable logic controllers (PLCs). The dataset includes measurements both in normal operation and under attack conditions, and label information is provided. The dataset contains $58$ features that represent sensor readings, actuator statuses, timing information, throughput time, and PLC parameters for the four stations. 
The processes covered include pick and place, loading, air pressing, controlling panels, sorting, and other related activities. The experimental setup is shown in Figure \ref{fig:DT_data}. In addition, we evaluate the proposed methods on the \emph{BATADAL} (Battle of the Attack Detection Algorithms) dataset \cite{Taormina2018BattleNetworks}, which provides realistic cyber-physical traces from a water distribution system subject to cyberattacks. The dataset simulates supervisory control and data acquisition (SCADA)-based IIoT environments where multiple pumping stations, water tanks, and sensors are monitored and controlled. Normal operating data are interspersed with malicious behaviors, including sensor spoofing, command injection, and control logic manipulation, designed to mimic real-world adversarial scenarios in industrial control systems.}

\subsection{Implementation Details}\label{sec:Implementation}

We utilized the PyTorch deep learning framework to implement the proposed digital twin-based learning mechanisms. A simple neural network model is used for both the digital twin and physical asset local models. The architecture consists of an input layer matching the feature size of the training data, two hidden layers with $16$ and and $8$-neurons with ReLU activation, and a single neuron output layer with sigmoid activation.
All models are trained by the Adam optimizer \cite{Kingma2014Adam:Optimization} with a mini-batch size of $32$, a learning rate of $0.001$. For FPF, we set the hyperparaeters $\beta=0.5$ and $\alpha=1$. For LPE, we set the layer exchange policy that sends the first hidden layer from clients to the digital twin and the second hidden layer from the digital twin to the clients. For DTKD, we pretrained the digital twin model for $5$ epochs to generate the pseudolabels required for training the client models. When implementing different methods, we consider scenarios with the following parameters selected: number of epochs $E=\{3,6,9\}$, batch size $B=\{10,20,30\}$, and fraction of active physical systems $C=\{0.3,0.6,0.9\}$. 
We simulate federated learning by partitioning the real-world dataset across multiple virtual assets and assumed $K=20$ systems.

\textcolor{black}{
In this study, we assume synchronous rounds and similar devices, but each method can scale to asynchronous and heterogeneous settings with minor extensions. FPF can weight client updates by both DT similarity and recency/reliability, so late or noisy contributions have less influence. LPE can become bandwidth-aware by exchanging only the most impactful layers per client budget and applying gentle, versioned merges to handle uneven links. CWA can alternate DT/client influence on fixed wall-clock intervals (instead of rounds), naturally tolerating stragglers and variable return times. DTML can add simple stability controls (e.g., proximal or control-variate style corrections) to keep meta-updates stable under non-IID data and uneven compute. DTKD can use lightweight teachers, cached/quantized soft targets, and reduced pull frequency for weak devices. Across the framework, partial participation, staleness caps, and basic compression (quantization/sparsification) provide practical robustness to varied bandwidth, compute, and participation levels.}

\subsection{Baseline and Evaluation Metrics} \label{sec:baseline}
We performed a comprehensive comparison of the proposed algorithms with a non-digital twin based standard Federated stochastic gradient descent (FedSGD), where each physical asset computes the gradient on its local data and the server aggregates those gradients to update the global model \cite{McMahan2017Communication-EfficientData}.
For performance analysis, the global model is evaluated on 20\% test data from the real-world dataset using various metrics, including accuracy ($A$), F1-score ($F_1$), precision ($P$), and recall ($R$).  
These metrics depend on the number of correctly-detected anomalies (true positives (TP)), the number of erroneously-detected anomalies (false positives (FP) or false alarms), the number of correctly-identified normal samples (true negatives (TN)), and the number of erroneously-identified normal samples (false negatives (FN)). 
By defining the true positive rate (TPR) and the false positive rate (FPR) as:
\begin{align}
    {\rm TPR}=\frac{{\rm TP}}{{\rm TP}+{\rm FN}} \;,\;
    {\rm FPR}=\frac{{\rm FP}}{{\rm FP}+{\rm TN}} \;,
\end{align}
The selected performance metrics are computed as follows:
\begin{align}
&A=\frac{\text{TP}+\text{TN}}{\text{TP}+\text{TN}+\text{FP}+\text{FN}}, \\
&P=\frac{{\rm TP}}{{\rm TP}+{\rm FP}}\;,\; 
R={\rm TPR}\;,\;
F_1=\frac{2PR}{P+R}\;.
\end{align}
Also, from the Receiver Operating Characteristic (ROC), the curve representing TPR vs. FPR (commonly used to evaluate models at different threshold values), we consider the Area under the ROC (AUC) as a relevant performance metric.



\section{Results and Discussions}\label{sec:Results} 

\subsection{Convergence Analysis}

\begin{figure*}
\centering
\begin{subfigure}{0.50\columnwidth}\centering
\includegraphics[height=0.75\columnwidth,width=\columnwidth]{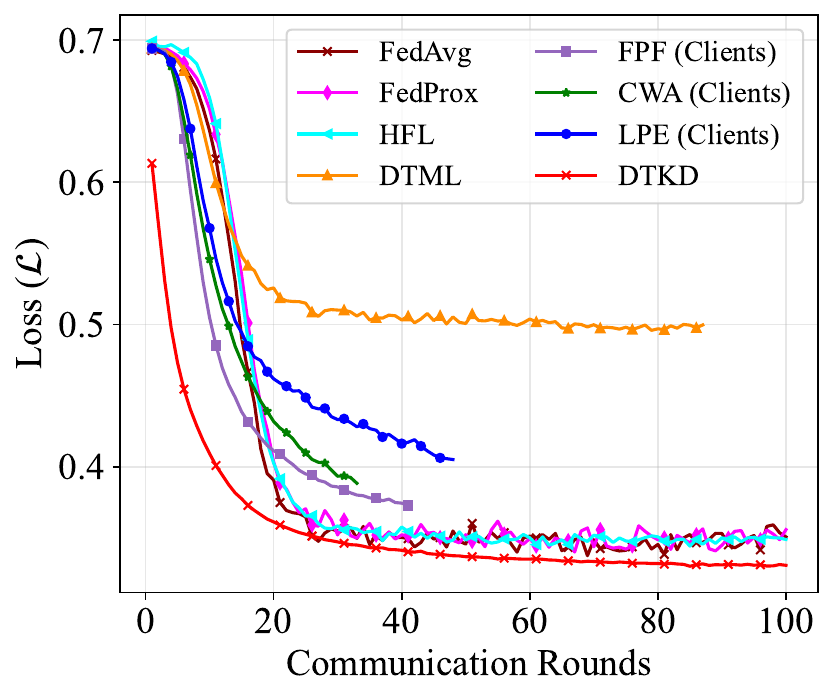}
\caption{Physical clients (I4.0)}
\end{subfigure} 
\begin{subfigure}{0.50\columnwidth}\centering
\includegraphics[height=0.75\columnwidth,width=\columnwidth]{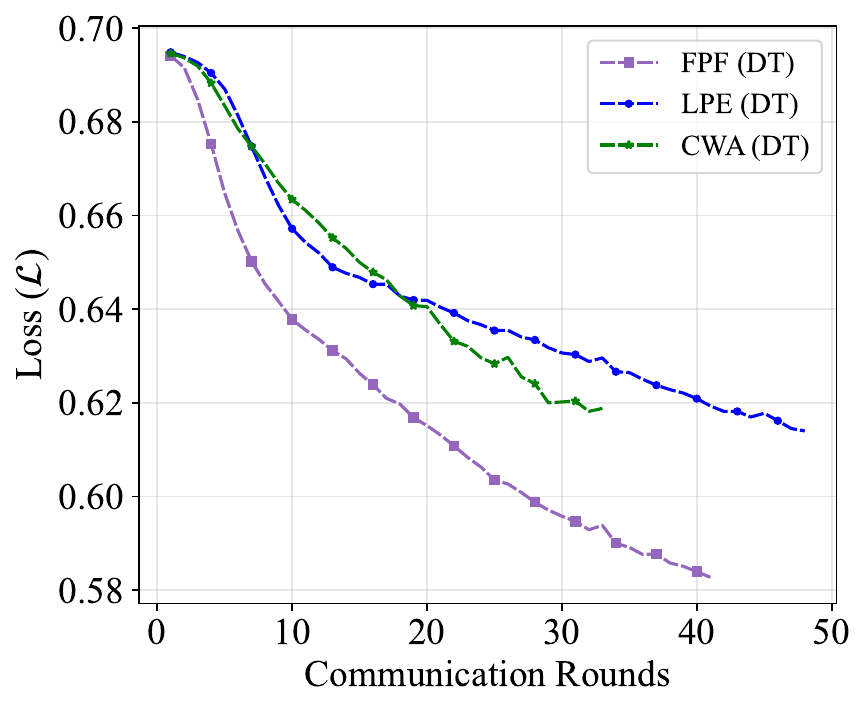}
\caption{Digital Twin (I4.0)}
\end{subfigure}
\begin{subfigure}{0.50\columnwidth}\centering
\includegraphics[height=0.75\columnwidth,width=\columnwidth]{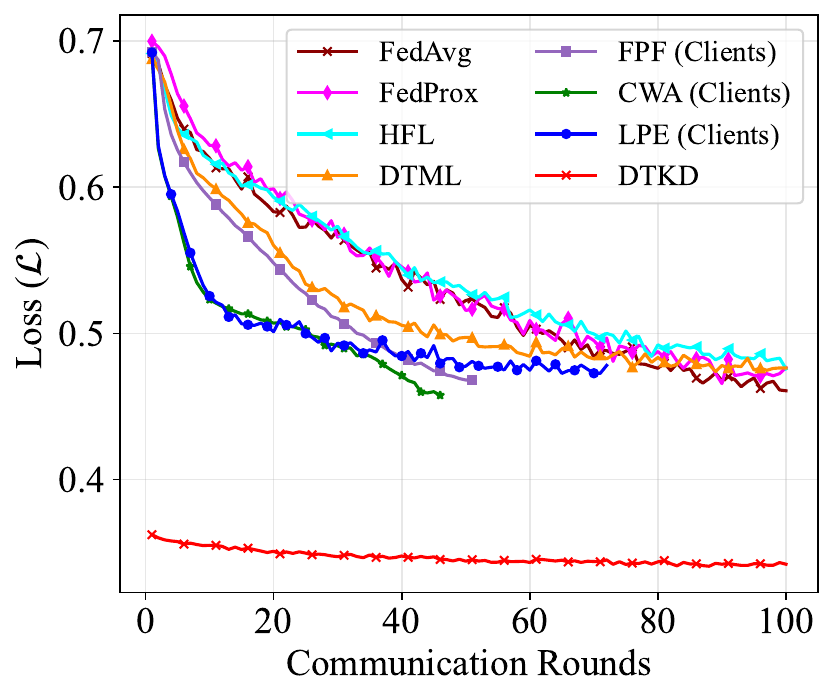}
\caption{Physical clients (BATADAL)}
\end{subfigure} 
\begin{subfigure}{0.50\columnwidth}\centering
\includegraphics[height=0.75\columnwidth,width=\columnwidth]{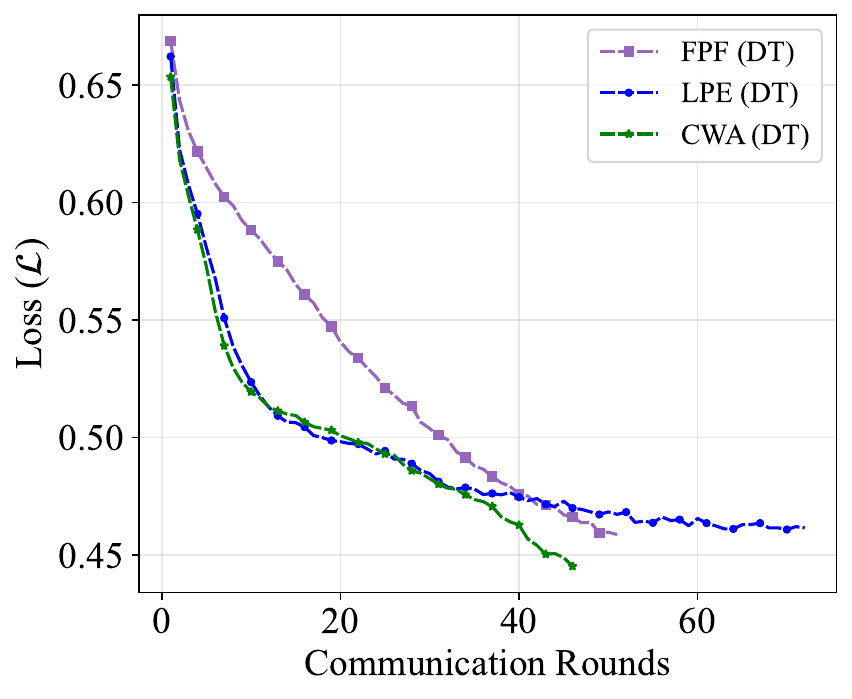}
\caption{Digital Twin (BATADAL)}
\end{subfigure}
\caption{Convergence behavior of DT-FL methods and baselines.}
\label{fig:convergence}
\end{figure*}
\textcolor{black}{
To evaluate the convergence behavior of the proposed digital twin-based federated learning methods, we conducted experiments under a fixed setting: total clients $K=20$, client fraction $C=0.3$, local batch size $B=10$, and local epochs $E=2$. The target anomaly detection accuracy was set to $80\%$, and the maximum number of communication rounds was capped at $100$ to prevent infinite loops.
Figure~\ref{fig:convergence} shows client- and DT-side convergence for both datasets. 
Among all methods, CWA converged fastest (33 rounds), as alternating twin–client updates enabled rapid synchronization. FPF followed (41 rounds), offering both speed and stability, while LPE reached convergence in 48 rounds, benefiting from selective cross-domain knowledge transfer. DTML required more rounds (87), reflecting its focus on long-term adaptability.  By contrast, FedAvg, DTKD, and FedProx failed to hit the 80\% target within 100 rounds. Although FedProx introduced a proximal term to stabilize heterogeneous updates, it could not fully exploit DT knowledge, limiting acceleration. Similarly, Hierarchical Learning (HL) converged faster than FedAvg but lagged behind DT-integrated methods, showing that multi-level aggregation alone is insufficient without explicit twin guidance.  Overall, DT-integrated approaches clearly outperform conventional FL baselines. CWA, FPF, and LPE deliver the best trade-offs between speed, robustness, and communication efficiency, making them attractive for real-time IIoT anomaly detection.}

\subsection{Anomaly Detection Performance}

\begin{figure*}
\centering
\begin{subfigure}{0.32\linewidth}\centering
\includegraphics[height=0.75\columnwidth,width=\columnwidth]{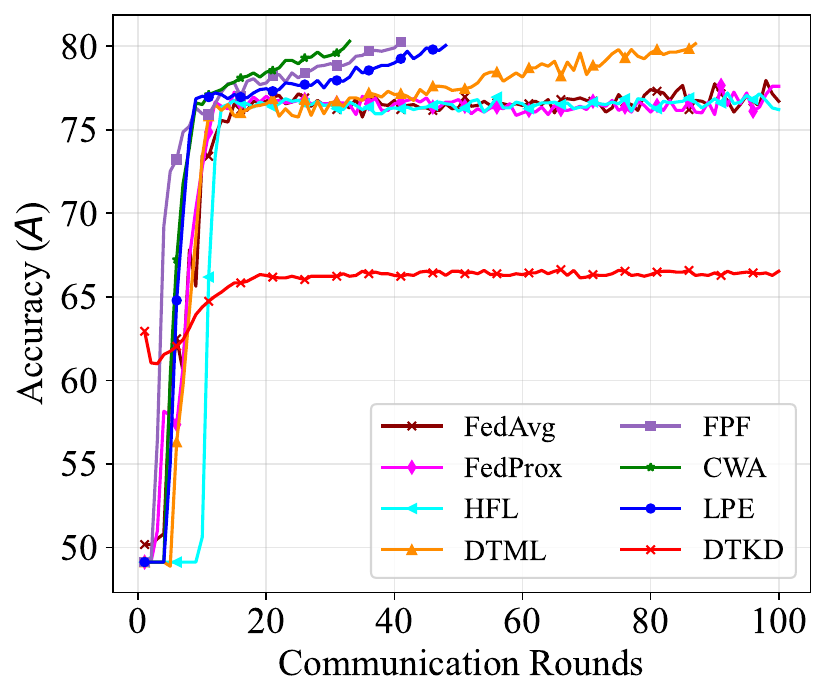}
\caption{Accuracy (I4.0)}
\end{subfigure} 
\begin{subfigure}{0.32\linewidth}\centering
\includegraphics[height=0.75\columnwidth,width=\columnwidth]{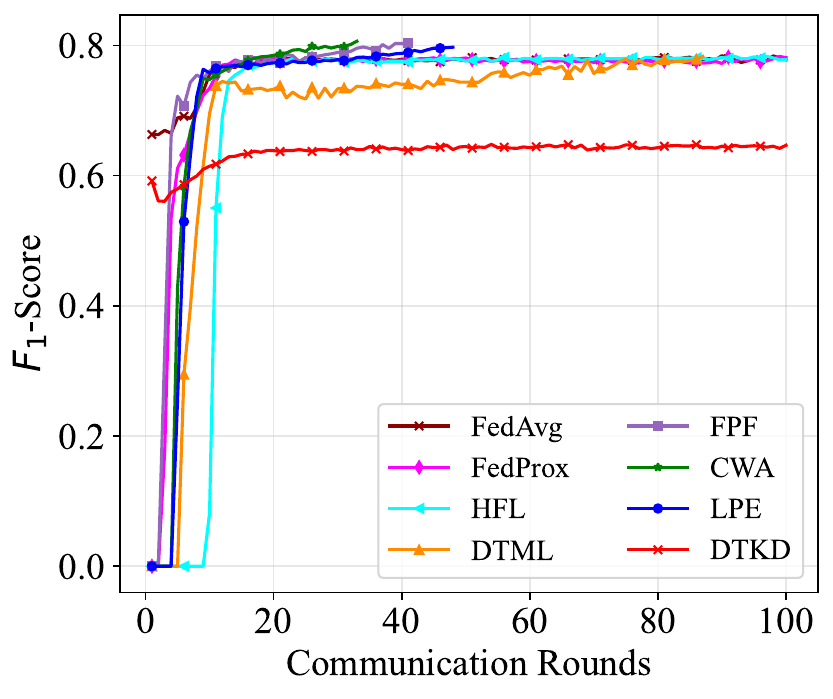}
\caption{$F_1$ (I4.0)}
\end{subfigure}
\begin{subfigure}{0.32\linewidth}\centering
\includegraphics[height=0.75\columnwidth,width=\columnwidth]{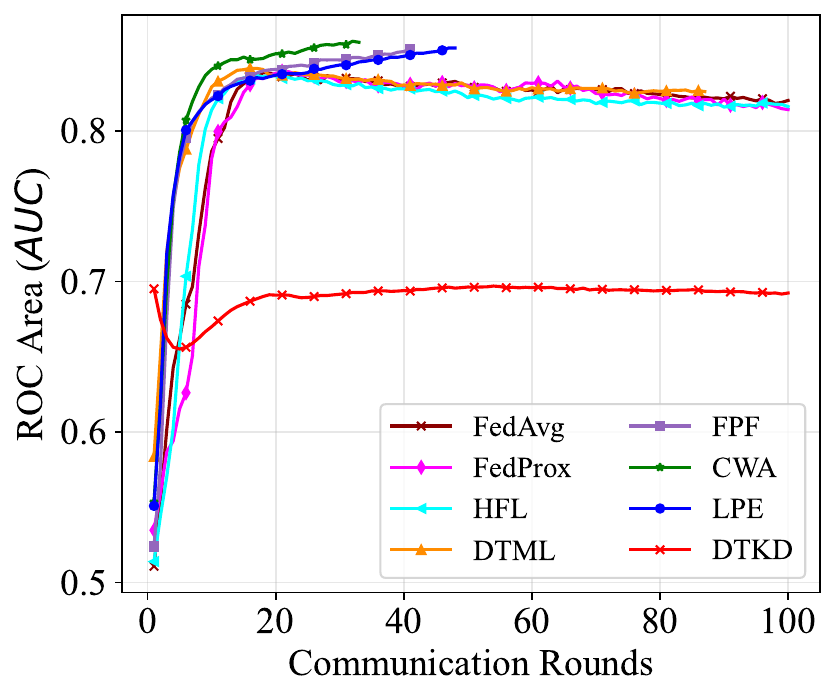}
\caption{AUC (I4.0)}
\end{subfigure}
\\
\begin{subfigure}{0.32\linewidth}\centering
\includegraphics[height=0.75\columnwidth,width=\columnwidth]{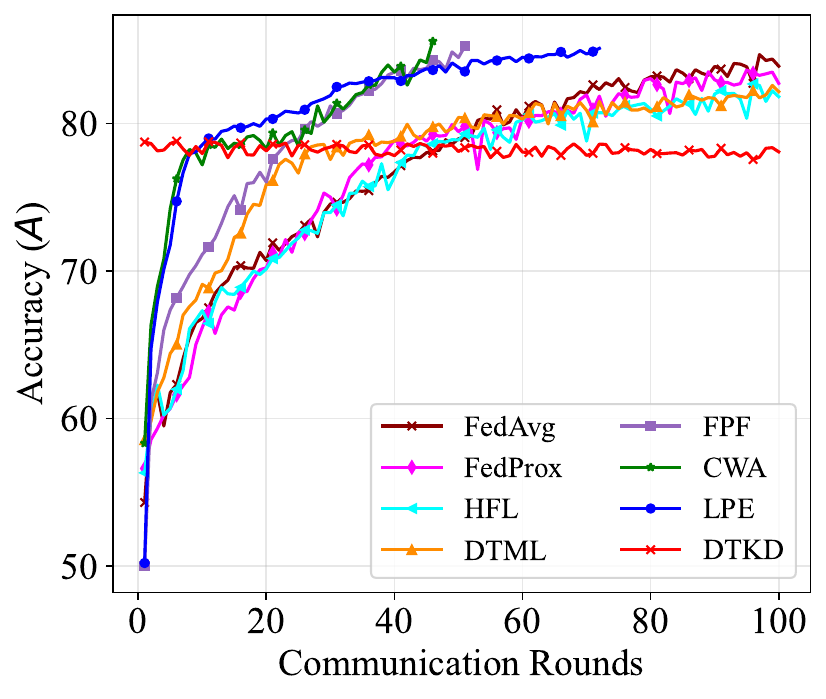}
\caption{Accuracy (BATADAL)}
\end{subfigure} 
\begin{subfigure}{0.32\linewidth}\centering
\includegraphics[height=0.75\columnwidth,width=\columnwidth]{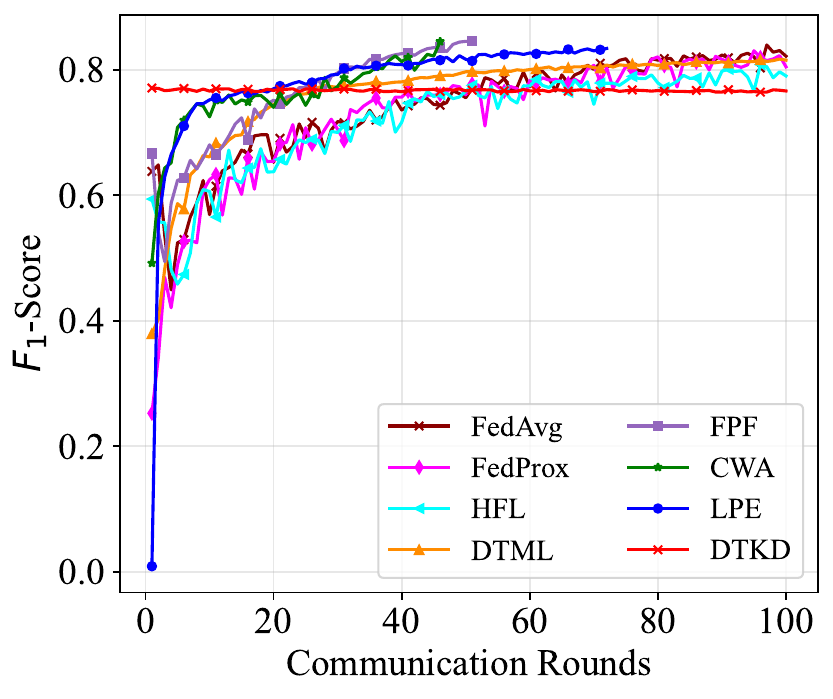}
\caption{$F_1$ (BATADAL)}
\end{subfigure}
\begin{subfigure}{0.32\linewidth}\centering
\includegraphics[height=0.75\columnwidth,width=\columnwidth]{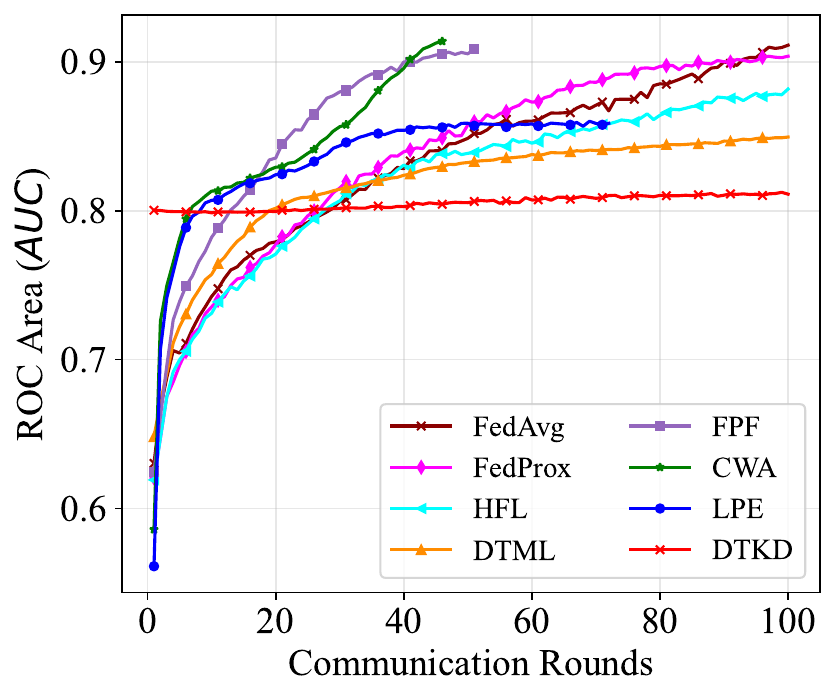}
\caption{AUC (BATADAL)}
\end{subfigure}
\caption{Detection performance across DT-FL methods and baselines.}
\label{fig:detection_performance}
\end{figure*}

\textcolor{black}{
We evaluate the anomaly detection performance of the global model using three metrics: accuracy, F1 score, and AUC. The experimental configuration is again fixed at $K=20$, $C=0.3$, $B=10$, and $E=2$, with a maximum of 100 communication rounds and a target accuracy of 80\%.
As shown in Figure \ref{fig:detection_performance}, for the I4.0 dataset, the baseline algorithms (FedAvg, FedProx and HFL) fail to achieve the target accuracy within the maximum communication round. CWA consistently outperformed other methods, reaching $80.3\%$ accuracy, $F_1$$\approx$0.81, and AUC $\approx$ 0.86 in only 33 rounds. FPF achieved 80.2\% accuracy, $F_1$ score of 0.80, AUC of 0.85 in 41 rounds, benefiting from vector-similarity weighting. LPE also performed strongly (accuracy=80.0\%, $F_1$=0.80, AUC=0.85), leveraging selective transfer across layers with moderate communication cost. DTML achieved competitive accuracy (80.1\%) but required nearly three times as many rounds (87), reflecting its emphasis on generalization under non-IID settings.  By contrast, DTKD underperformed (66.5\% accuracy, $F_1=0.65$, AUC=0.69), confirming the limitations of one-way distillation from a static DT teacher. FedAvg plateaued at 76.7\% accuracy, while FedProx showed slightly improved stability under heterogeneity but failed to surpass 77.5\%. The trends were consistent across both datasets: DT-FL methods, particularly CWA and FPF, achieved faster convergence, higher accuracy, and stronger generalization than standard FL baselines. These results highlight the advantage of explicitly coupling digital twin knowledge with client updates for real-world IIoT anomaly detection.  
}

\subsection{Parameter Sensitivity Analysis}

\begin{table*}[h!]
\centering
\renewcommand{\arraystretch}{1.5}
\begin{tabular}{>{\color{black}}c>{\color{black}}c>{\color{black}}c>{\color{black}}c>{\color{black}}c>{\color{black}}c>{\color{black}}c>{\color{black}}c>{\color{black}}c>{\color{black}}c}
\hline
\multirow{2}{*}{\textbf{Parameters}} & & \multicolumn{2}{c}{\textbf{DTML}} & \multicolumn{2}{c}{\textbf{FPF}} & \multicolumn{2}{c}{\textbf{CWA}} & \multicolumn{2}{c}{\textbf{LPE}}\\
\cline{3-10}
 & & F1 (\%) & AUC (\%) & F1 (\%) & AUC (\%) & F1 (\%) & AUC (\%) & F1 (\%) & AUC (\%) \\
\hline
\multirow{3}{*}{E} 
 & 3 & 78.32 & 82.60 & 80.34 & 85.45 & 80.65 & 85.88 & 79.74 & 85.50 \\
 & 6 & 78.21 & 83.58 & 80.61 & 84.45 & 80.58 & 84.26 & 79.78 & 88.87 \\
 & 9 & 78.33 & 85.17 & 80.66 & 84.54 & 80.51 & 82.48 & 79.82 & 87.88 \\
\hline
\multirow{3}{*}{B} 
 & 10 & 78.32 & 82.60 & 80.34 & 85.45 & 80.65 & 85.88 & 79.74 & 85.50 \\
 & 20 & 78.22 & 84.35 & 80.18 & 85.90 & 80.54 & 85.68 & 79.94 & 86.55 \\
 & 30 & 76.44 & 85.48 & 78.66 & 83.89 & 80.57 & 84.10 & 80.71 & 85.53 \\
\hline
\multirow{3}{*}{C} 
 & 0.3 & 78.32 & 82.60 & 80.34 & 85.45 & 80.65 & 85.88 & 79.74 & 85.50 \\
 & 0.6 & 77.64 & 82.42 & 80.34 & 85.45 & 80.65 & 85.88 & 79.74 & 85.50 \\
 & 0.9 & 78.19 & 82.80 & 80.34 & 85.45 & 80.65 & 85.88 & 79.74 & 85.50 \\
\hline
\end{tabular}
\caption{\textcolor{black}{Performance metrics (F1-score and AUC expressed in percentages) for different parameter settings across DTML, FPF, CWA, and LPE methods.}}
\label{tab:param_sensitivity}
\end{table*}

\begin{figure*}
\centering
\begin{subfigure}{0.65\columnwidth}\centering
\includegraphics[height=0.7\columnwidth,width=\columnwidth]{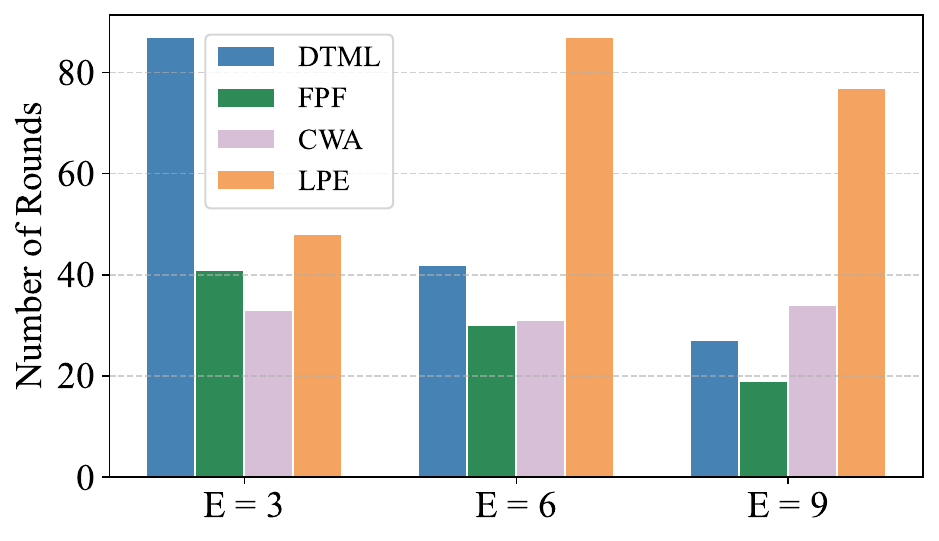}
\caption{Local epoch}
\label{fig:comparison_E}
\end{subfigure} 
\begin{subfigure}{0.65\columnwidth}\centering
\includegraphics[height=0.7\columnwidth,width=\columnwidth]{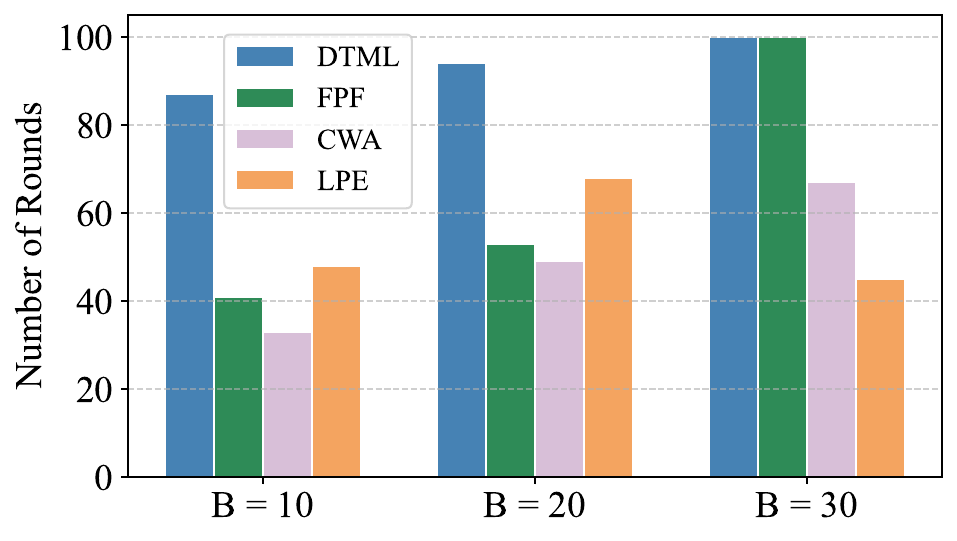}
\caption{Local batch}
\label{fig:comparison_B}
\end{subfigure}
\begin{subfigure}{0.65\columnwidth}\centering
\includegraphics[height=0.7\columnwidth,width=\columnwidth]{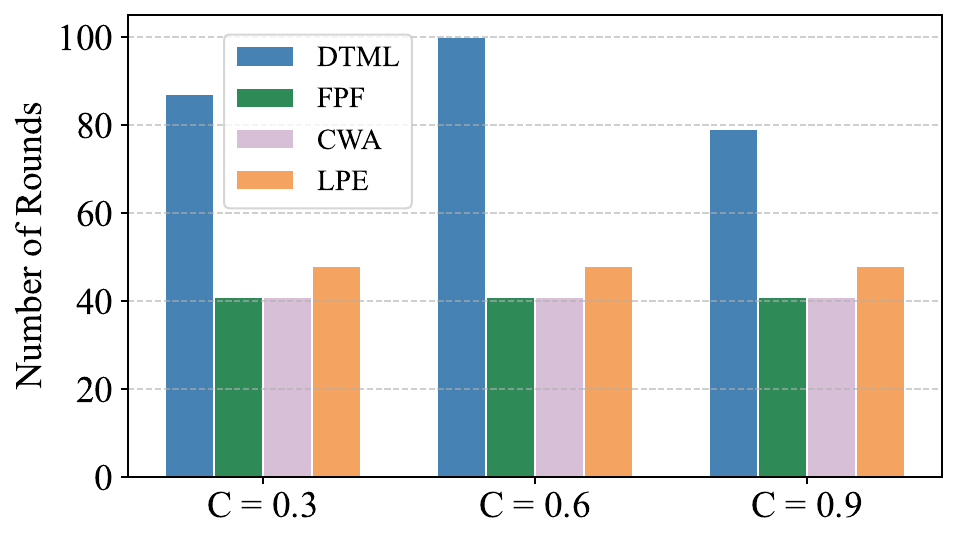}
\caption{Client fraction}
\label{fig:comparison_C}
\end{subfigure}
\caption{Parameter sensitivity analysis}
\label{fig:parameter_sensitivity}
\end{figure*}

\color{black}

To investigate the robustness of the proposed digital twin-based federated learning methods, we conduct a parameter sensitivity analysis over three key hyperparameters: the number of local epochs ($E$), local batch size ($B$), and client fraction ($C$). The results are summarized in 
Table~\ref{tab:param_sensitivity} and Figure~\ref{fig:parameter_sensitivity}).  

\subsubsection{Local Epochs ($E$)} Increasing $E$ improved convergence for all methods up to a moderate level. FPF achieved the fastest convergence at $E=9$ (19 rounds, F1=0.81), while CWA and LPE remained stable with strong F1 and AUC. LPE reached the highest AUC (88.9\%) at $E=6$, highlighting its discriminative strength.  

\subsubsection{Batch Size ($B$)} Small batches ($B=10$) yielded the best trade-off between stability and convergence (e.g., FPF: 41 rounds, F1=0.80, AUC=0.85). At larger $B$, DTML and FPF degraded, while LPE remained robust, benefiting from selective layer exchange that mitigates overfitting to biased gradients.  

\subsubsection{Client Fraction ($C$)} DTML was most sensitive to higher client fractions, with longer convergence despite stable AUC. FPF maintained steady convergence (41 rounds) across all $C$, while CWA and LPE consistently achieved target accuracy in 33–48 rounds with F1 $\approx$0.81 and AUC $>$0.85, demonstrating strong scalability.  

In general, FPF provides the fastest and most consistent convergence, CWA balances speed and generalization, and LPE excels under large batches and diverse client settings. DTML adapts well in low-heterogeneity cases but is more sensitive to high client diversity. These findings confirm that DTFL methods remain robust across parameter variations, with each offering unique strengths for IIoT deployment.  

\color{black}

\subsection{Ablation Study}
\subsubsection{FPF}

\begin{figure}[t!]\centering
\includegraphics[width=0.8\columnwidth]{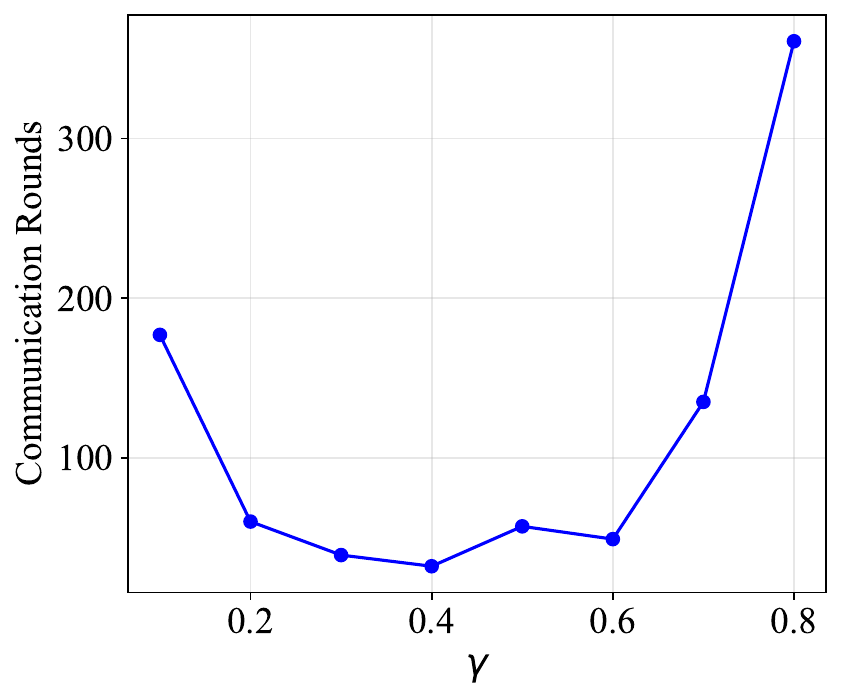}
\caption{\textcolor{black}{FPF Weighting factor ($\gamma$) vs. minimum communication round to reach 80\% accuracy}}
\label{fig:gamma_vs_rounds}
\end{figure}

To assess the impact of the fusion hyperparameter $\gamma$ in FPF, we conducted a sensitivity analysis by varying $\gamma$ from 0.1 to 0.9 and recording the minimum number of communication rounds required to achieve $80\%$ accuracy. The results, shown in Fig.~\ref{fig:gamma_vs_rounds}, reveal a non-linear relationship between $\gamma$ and convergence speed. Optimal performance was observed for $\gamma=0.3$ and $\gamma=0.4$, which reached the target accuracy in only 39 and 32 rounds, respectively. In contrast, very small ($\gamma=0.1$) or very large ($\gamma=0.8$, $\gamma=0.9$) values significantly slowed convergence, requiring 177 and 361 rounds, with the model failing to reach the target accuracy for $\gamma=0.9$ within the 500-round limit. These findings suggest that balanced weighting between the digital twin and client models is crucial: excessively favoring either side degrades convergence, while moderate fusion provides the best trade-off between accuracy and efficiency.  

In addition to fixed $\gamma$, we also investigated adaptive weighting strategies based on matrix similarity measures, including cosine similarity, RV coefficient, and mutual information. The number of rounds required to reach $80\%$ accuracy was 202 for cosine similarity, 78 for the RV coefficient, and 282 for mutual information. Among these, the RV coefficient provided the fastest convergence, confirming it as a more effective similarity measure for this setting. Nevertheless, adaptive similarity measures are particularly valuable in scenarios involving adversarial or highly heterogeneous clients, whereas a fixed grid search over $\gamma$ values is more stable and effective in the current context.  

\subsubsection{LPE}
To investigate the effect of the Layer-wise Parameter Exchange (LPE) strategy, we compared the baseline static policy, where lower layers are exchanged from the digital twin and upper layers from clients, with its reverse policy counterpart, in which lower layers are taken from clients and upper layers from the digital twin. The results indicate that the baseline policy achieves the target accuracy of 80\% within 50-60 communication rounds, whereas the reverse policy only reaches 75.71\% even after the maximum of 100 communication rounds. This performance gap highlights that lower layers capture more generalizable feature representations across domains, making them better suited for sharing from the digital twin, while upper layers tend to be task-specific and should be adapted from client data. These findings validate the design choice in the baseline LPE approach and emphasize the importance of exchange direction in achieving faster convergence and higher accuracy. 

\subsubsection{DTKD}

\begin{figure}[t!]\centering
\includegraphics[width=0.8\columnwidth]{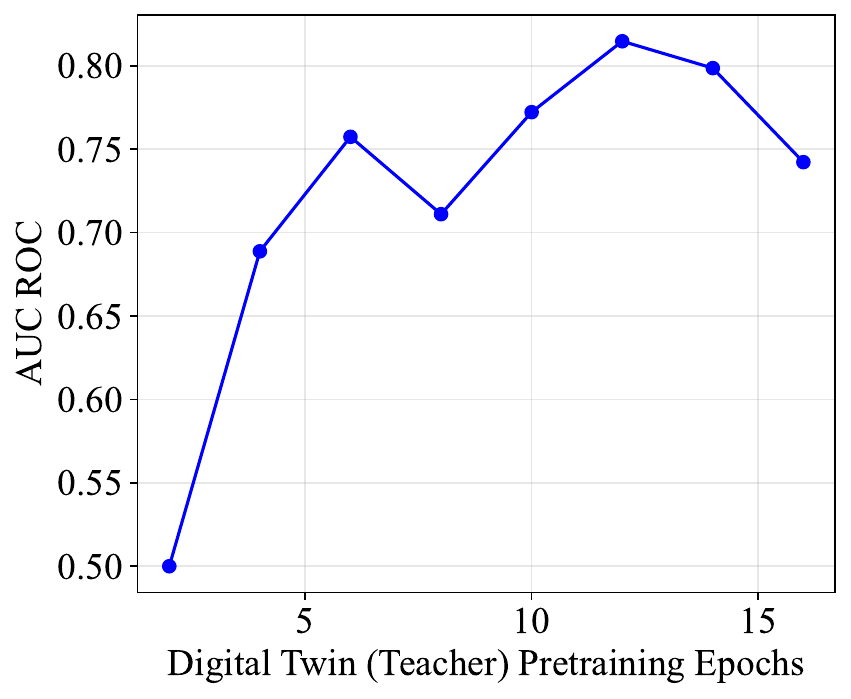}
\caption{\textcolor{black}{DTKD Sensitivity: Teacher Pretraining Epochs vs AUC-ROC}}
\label{fig:DTKD_epoch_vs_ROC}
\end{figure}

To determine the risk of bias or overfitting in the teacher DT training stage, we analyze DT pretraining epochs with the global model’s AUC-ROC. As shown in  Figure \ref{fig:DTKD_epoch_vs_ROC}, the curve is non-monotonic, i.e., AUC improves as the teacher is trained for a moderate number of epochs (mid-range) and then plateaus or slightly declines with further teacher training. This pattern suggests a classic bias–variance trade-off at the teacher level: (i) with too few epochs, the teacher underfits and provides noisy/low-signal soft targets; (ii) with excessive training, the teacher overfits to synthetic DT data, producing overconfident, low-entropy targets that do not transfer well to heterogeneous client distributions, which harms student generalization. The best distillation signal arises when the teacher is strong but still well-calibrated, i.e., around the mid-range of pretraining.

\color{black}

\subsection{Scalability Analysis}
\label{subsec:scalability_results}

\begin{figure}\centering
\begin{subfigure}{0.93\columnwidth}\centering
\includegraphics[height=0.75\columnwidth,width=\columnwidth]{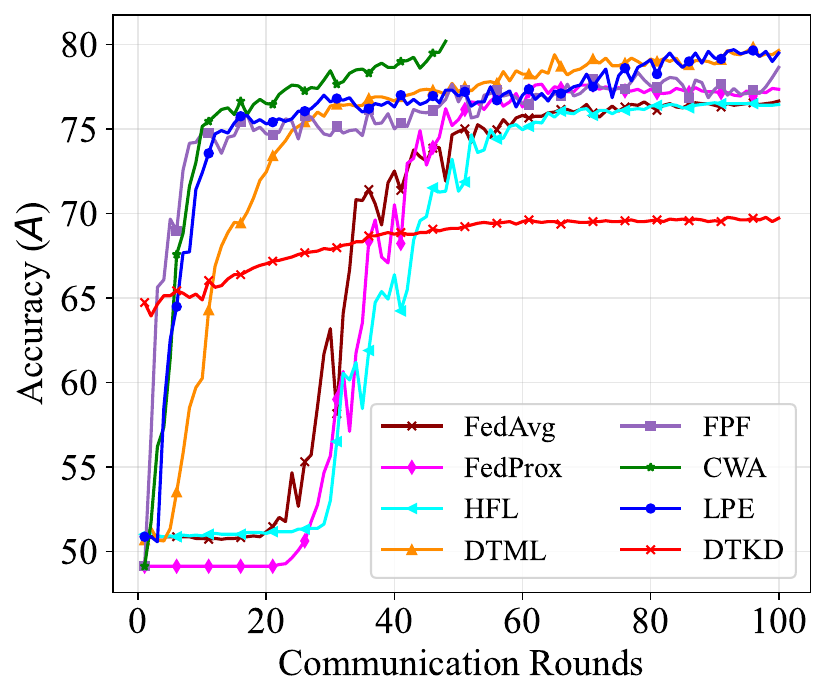}
\caption{\textcolor{black}{Accuracy}}
\label{fig:accuracy_fixed_all_k_100}
\end{subfigure} 
\begin{subfigure}{0.93\columnwidth}\centering
\includegraphics[height=0.75\columnwidth,width=\columnwidth]{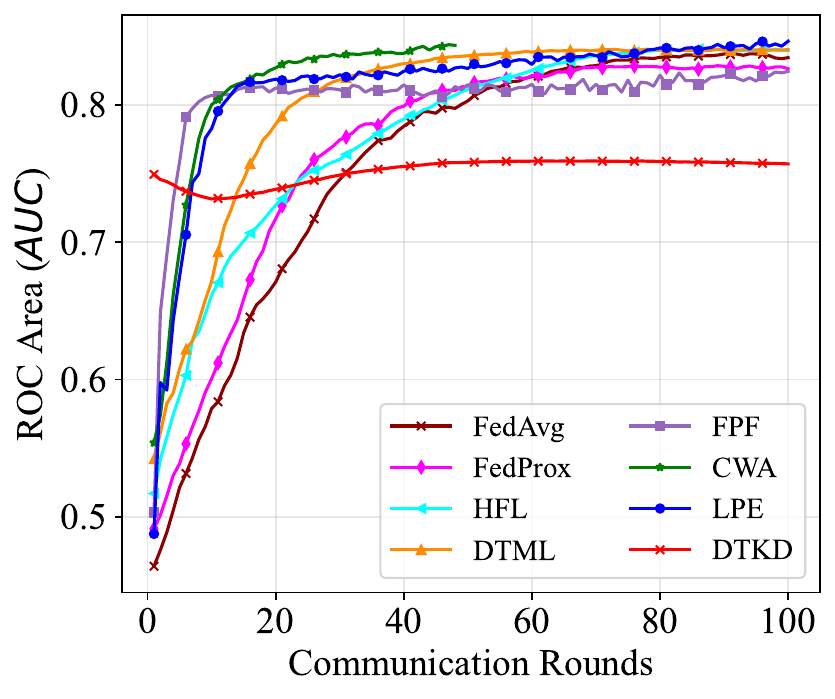}
\caption{\textcolor{black}{ROC-AUC}}
\label{fig:auc_fixed_all_k_100}
\end{subfigure}
\caption{\textcolor{black}{Scalability Analysis for K=100.}}
\label{fig:Scalability}
\end{figure}

To assess scalability, we extended the experiments to $K=100$ clients with a maximum limit of 100 communication rounds, targeting $80\%$ accuracy as the stopping criterion. The results are shown in Figure ~\ref{fig:Scalability}, where detection accuracy and AUC are plotted against the number of rounds. As illustrated in Figure~\ref{fig:accuracy_fixed_all_k_100}, CWA consistently achieved the fastest convergence, reaching the $80\%$ accuracy threshold well within the 100-round limit. Baselines such as FedAvg, FedProx, and HFL performed consistently worse than the proposed DTFL methods, reinforcing their scalability advantage. DTKD remained the weakest performer, unable to reach the target accuracy within the budget. Overall, the $K=100$ experiments demonstrate that our DTFL methods retain strong detection performance and convergence properties in large-scale IIoT deployments, supporting their suitability for scenarios with hundreds of clients.

\subsection{Distributional Alignment}
\label{subsec:dist-align}
\begin{figure}\centering
\begin{subfigure}{0.93\columnwidth}\centering
\includegraphics[height=0.75\columnwidth,width=\columnwidth]{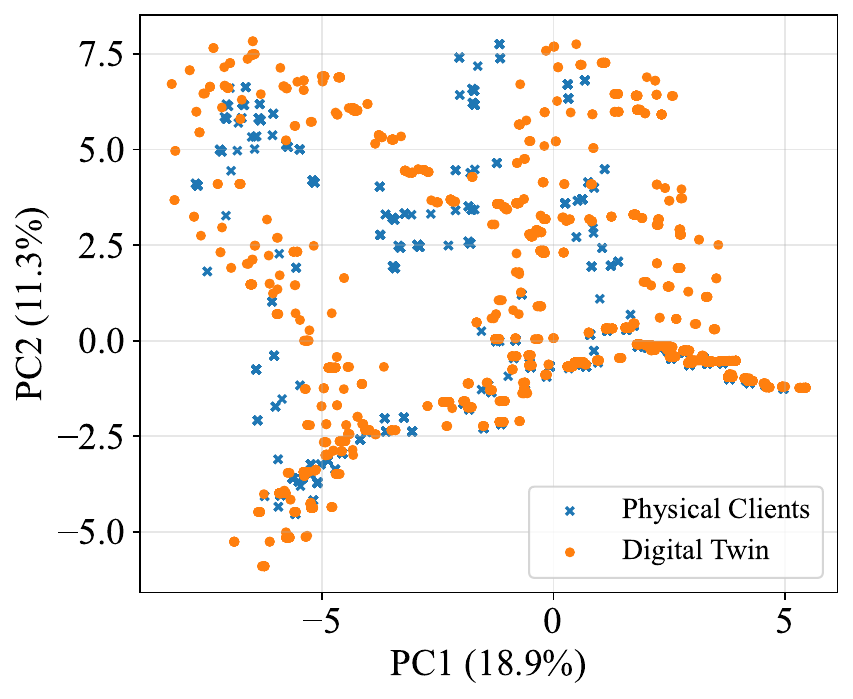}
\caption{\textcolor{black}{I4.0}}
\label{fig:PCA_data_1}
\end{subfigure} 
\begin{subfigure}{0.93\columnwidth}\centering
\includegraphics[height=0.75\columnwidth,width=\columnwidth]{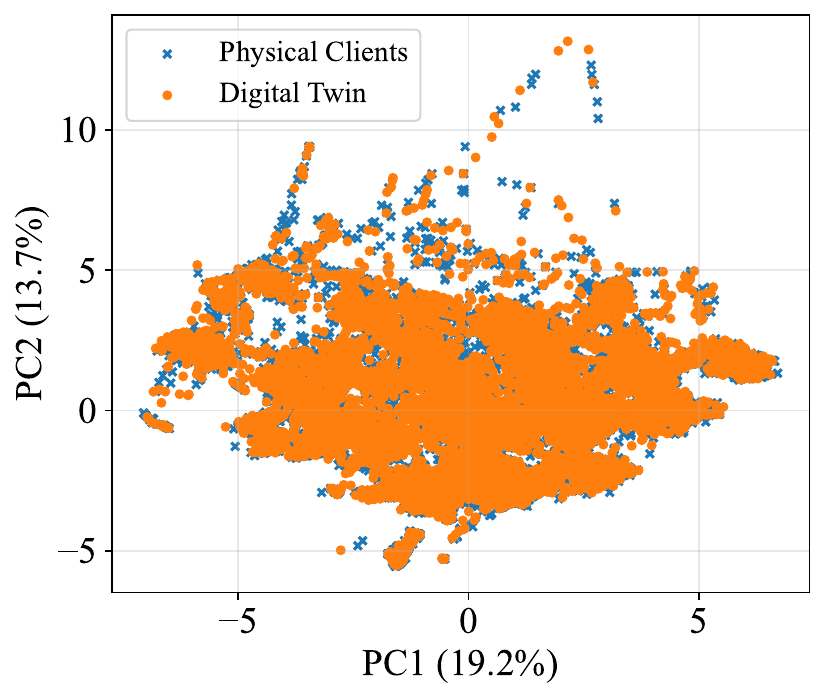}
\caption{\textcolor{black}{BATADAL}}
\label{fig:PCA_data_2}
\end{subfigure}
\caption{\textcolor{black}{Distributional alignment via PCA.}}
\label{fig:PCA_data}
\end{figure}

To quantify how well the digital-twin data aligns with the real data, we jointly standardize features and compute: (i) the average absolute mean gap $\mathbf{\overline{|\Delta\mu|}}=\frac{1}{d}\sum_{j=1}^d |\mu^{\text{real}}_j-\mu^{\text{dt}}_j|$, (ii) the average absolute variance gap $\mathbf{\overline{|\Delta\sigma^2|}}=\frac{1}{d}\sum_{j=1}^d |\sigma^{2,\text{real}}_j-\sigma^{2,\text{dt}}_j|$, (iii) the linear maximum mean discrepancy (MMD), 
and (iv) the sliced Wasserstein distance (SWD). Lower values indicate better alignment. Table \ref{tab:dist-align} summarizes distributional alignment between physical and digital twin data.
\begin{table}[t]
\centering
\caption{Distributional alignment between Real and DT data.}
\label{tab:dist-align}
\begin{tabular}{>{\color{black}}l>{\color{black}}c>{\color{black}}c}
\toprule
\textbf{Metric} & \textbf{I4.0} & \textbf{BATADAL} \\
\midrule
\textbf{Samples (Real/DT)} & 5152 / 4867 & 12446 / 12446 \\
\textbf{Features} & 57 & 36 \\
$\mathbf{\overline{|\Delta\mu|}}$ & 0.1655 & 0.0096 \\
$\mathbf{\overline{|\Delta\sigma^2|}}$ & 0.3062 & 0.0110 \\
\textbf{MMD} & 2.6230 & 0.0711 \\
\textbf{SWD} & 0.3931 & 0.0230 \\
\bottomrule
\end{tabular}
\vspace{2mm}\\
\end{table}
We additionally plot PCA (2D) on the jointly standardized pooled data (Real vs.\ DT) to visualize overlap as shown in Figure \ref{fig:PCA_data}.
For the cyber-physical dataset, the larger mean/variance gaps and higher MMD/SWD are reflected by clearer separation between real and DT point clouds in the first two PCs (centroids shifted with only partial overlap), indicating a moderate covariate shift that justifies DT to Real alignment mechanisms (e.g., fusion, layer-wise exchange).
Conversely, for BATADAL the PCA scatter shows strong overlap between Real and DT clusters (minor centroid shift, comparable spread), consistent with the near-zero moment gaps and small MMD/SWD, suggesting good distributional alignment.

\subsection{Computational Complexity}

\begin{table*}[!t]
\centering
\begin{tabular}{@{}>{\color{black}}p{1cm}>{\color{black}}p{4cm}>{\color{black}}p{3.5cm}>{\color{black}}p{3cm}>{\color{black}}p{2cm}@{}}
\toprule
\textbf{Method} & \textbf{Client Time} & \textbf{Server Time} & \textbf{Communication (Up/Down)} & \textbf{Extra Memory (Server/Client)} \\
\midrule
FedAvg & $\mathcal{O}\big(E\,\frac{n_k}{B}\,P\big)$ & $\mathcal{O}(K P)$ & $\mathcal{O}(P) / \mathcal{O}(P)$ & None \\
FedProx & Same as FedAvg $+ \mathcal{O}(E\,\tfrac{n_k}{B}\,P)$ & $\mathcal{O}(K P)$ & Same as FedAvg & None \\
HFL & Same as FedAvg & $\mathcal{O}(H P) + \mathcal{O}(K P)$  & $\mathcal{O}(P) / \mathcal{O}(P)$ per tier & None \\
DTML & Same as FedAvg & $\mathcal{O}(K P) + \mathcal{O}(n_{\text{twin}} P)$ & Same as FedAvg & $\mathcal{O}(P)$/None \\
FPF & Same as FedAvg & $\mathcal{O}(K P) + \mathcal{O}(K)$ & Same as FedAvg & None \\
LPE & Same as FedAvg & $\mathcal{O}(K P)$ (copy only) & $\mathcal{O}(\sum_{\ell \in \mathcal{L}_{\text{tx}}} P_\ell)$ / same & None \\
CWA & Same as FedAvg & $\mathcal{O}(K P)$ & $2\mathcal{O}(P) / 2\mathcal{O}(P)$ & None \\
DTKD & $\mathcal{O}\big(E\,\tfrac{n_k}{B}\,P + n_k c\big)$ & $\mathcal{O}(n_{\text{twin}} P)$ & $\mathcal{O}(P)+\mathcal{O}(n_k c) / \mathcal{O}(P)$ & $\mathcal{O}(n_k c)$ \\
\bottomrule
\end{tabular}
\caption{\textcolor{black}{Computational complexity analysis (per FL round). $H$ is the number of edge aggregators in hierarchical FL.}}
\label{tab:complexity}
\end{table*}

We analyze the per-round complexity of each method across three stages: global model broadcast, local training, and aggregation. The dominant client cost for all methods remains $\mathcal{O}(E\,\tfrac{n_k}{B}\,P)$ from mini-batch SGD.  

\textbf{Baselines:} FedAvg serves as the reference. FedProx adds a proximal regularizer, incurring negligible overhead beyond FedAvg. Hierarchical FL introduces an intermediate aggregation layer with $H$ edge aggregators, leading to $\mathcal{O}(H P)$ additional server cost but similar communication per tier.  

\textbf{DTFL methods:} DTML adds $\mathcal{O}(n_{\text{twin}} P)$ for meta-updates on DT data. FPF incurs only $\mathcal{O}(K)$ extra similarity computations. LPE reduces communication to exchanged layers $\sum_{\ell \in \mathcal{L}_{\text{tx}}} P_\ell$. CWA doubles communication since both DT and client models are exchanged across rounds. DTKD introduces the largest client overhead due to teacher–student distillation, with extra $\mathcal{O}(n_k c)$ operations and memory for soft labels.  

In summary, FedProx and FPF are nearly cost-free modifications of FedAvg, LPE provides the best asymptotic communication savings, while DTKD is the most resource-intensive. Table~\ref{tab:complexity} compares all methods quantitatively.

\section{Conclusions and Future Work}\label{sec:Conclusions}

\textcolor{black}{
In this paper, we proposed a hybrid and communication-efficient anomaly detection framework that integrates digital twins with federated learning mechanisms to address key IIoT challenges, including data scarcity, privacy, heterogeneity, and communication overhead. Five methods were introduced: DTML, FPF, LPE, CWA, and DTKD, each enabling distinct mechanisms for combining synthetic DT knowledge with client data.  Extensive experiments demonstrated that CWA achieved the fastest convergence, while FPF provided the best trade-off between accuracy and generalization. LPE showed robustness under varying client settings, and sensitivity analyses confirmed the stability of all methods across hyperparameter variations. Overall, adaptive or bidirectional knowledge transfer strategies (e.g., FPF and CWA) consistently outperformed static approaches.  In general, integrating DTs into FL significantly enhances efficiency, robustness, and accuracy in IIoT anomaly detection. Future work will extend this study by: (i) investigating asynchronous and heterogeneous FL, (ii) developing adaptive LPE policies, (iii) designing memory/latency–aware DTKD via quantization and lightweight distillation, (iv) incorporating uncertainty estimation for safety-critical deployment, (v) mitigating digital twin synchronization delays and communication variability, (vi) introducing lightweight synchronization and hardware-aware scaling, and (vii) validating the framework on broader datasets and attack types.}

\color{black}

\bibliographystyle{IEEEtran}
\bibliography{references}

\end{document}